\documentclass[letterpaper]{article} 
\usepackage[preprint]{aaai2027}  
\usepackage[hyphens]{url}  
\usepackage{graphicx} 
\usepackage{natbib}  
\usepackage{caption} 
\usepackage{algorithm}
\usepackage{algorithmic}

\usepackage{newfloat}
\usepackage{listings}
\DeclareCaptionStyle{ruled}{labelfont=normalfont,labelsep=colon,strut=off} 
\floatstyle{ruled}
\newfloat{listing}{tb}{lst}{}
\floatname{listing}{Listing}

\usepackage{booktabs}

\usepackage{array}
\usepackage{xspace}
\usepackage{pifont}
\usepackage{placeins}
\usepackage{fancyvrb}
\usepackage{fvextra}
\usepackage[capitalize,noabbrev]{cleveref}
\newcommand{\sys}{KernelClaw\xspace}
\usepackage{multirow}

\title{\sys{}: A Multi-Agent Framework for GPU Kernel Optimization with Dual-Memory Mechanisms}
\author {
    Qitong Sun\textsuperscript{\rm 1,\rm 2},
    Jun Han\textsuperscript{\rm 1},
    Tianlin Li\textsuperscript{\rm 1,\rm 2}\corresponding,
    Zhe Tang\textsuperscript{\rm 2},
    Sheng Chen\textsuperscript{\rm 2},
    Fei Yang\textsuperscript{\rm 2},
    Aishan Liu\textsuperscript{\rm 1},
    Xianglong Liu\textsuperscript{\rm 1},
    Yang Liu\textsuperscript{\rm 2,\rm 3}
}
\affiliations {
    \textsuperscript{\rm 1}School of Computer Science and Engineering, Beihang University, China\\
    \textsuperscript{\rm 2}Zhejiang Lab, China\\
    \textsuperscript{\rm 3}Nanyang Technological University, Singapore\\
    tianlin001@buaa.edu.cn
}

\begin{document}

\maketitle

\begin{abstract}
Improving GPU kernel efficiency is crucial for efficient AI systems. Recent work has explored leveraging large language models (LLMs) for GPU kernel generation and optimization. 
However, existing LLM-based kernel optimization pipelines typically rely on opaque, implicitly learned heuristics within the LLMs to determine optimization strategies. This leads to inefficient trial-and-error and weakly interpretable optimizations. Our key insight is to replace implicit heuristics with expert optimization knowledge that is aware of task trajectories. 
Specifically, we present \sys{}, a multi-agent framework with a dual-level memory architecture.
\sys{} operates by coordinating agents with long-term memory of reusable expert knowledge and short-term memory to prevent repetitive backtracking.
On KernelBench Levels 1–3, \sys{} achieves a 100\% success rate and average speedups of 6.98×, 7.89×, and 1.94× over Torch Eager on Levels 1, 2, and 3, respectively, outperforming prior baselines. Code is available at \url{https://github.com/0satan0/KernelMem/}.
\end{abstract}


\section{Introduction}

Efficient GPU kernels are fundamental to modern AI systems, as they largely determine the throughput of core operators such as GEMM, convolution, and data movement \citep{paszke2019pytorch,tay2022efficienttransformerssurvey,pandey2022transformational,ansel2024pytorch}.
Achieving high performance, however, remains a highly expertise-intensive and architecture-dependent process, requiring careful coordination of memory access patterns, parallel decomposition, and synchronization. 
As a result, manual kernel optimization often demands lengthy profiling-guided iteration cycles, bottleneck diagnosis, targeted code refinement, and repeated validation, making the development process both costly and time-consuming.

Recent advances in large language models (LLMs) have opened a promising avenue for automated GPU kernel generation and optimization. Existing LLM-based kernel optimization approaches typically follow two paradigms: (i) training-based methods that collect data and train models for kernel optimization \citep{li2025cudal1improvingcudaoptimization,baronio2025kevinmultiturnrlgenerating}. However, such approaches require curating large-scale datasets and substantial training compute, leading to high cost and long training cycles, and often exhibit limited generalization capability.
To mitigate these limitations, agentic methods have been proposed, in which models iteratively refine kernels using compilation and correctness checks, together with profiling feedback \citep{zhang2025cudaforgeagentframeworkhardware,wei2025astramultiagentgpukernel,lei2025pragmaprofilingreasonedmultiagentframework,dong2025starkstrategicteamagents}. Although these agentic approaches have achieved commendable progress, important gaps in kernel optimization still remain. In particular, \ding{182} optimization decisions are often opaque, emerging from implicitly learned heuristics within LLMs that are weakly interpretable. 
This makes it difficult to understand why a given strategy is selected and, in turn, hinders effective correction and further improvement of the optimization process. In addition, \ding{183} these methods typically rely on multi-round iterative refinement, which can be redundant, i.e., repeatedly revisiting previously seen and incorrect optimization strategies, thereby increasing runtime overhead.

To address the above issues, we design two complementary mechanisms. First, we replace implicit, model-internal heuristics with explicit kernel optimization expertise. This expertise encodes reusable insights accumulated across kernels.
Second, we store task-specific optimization trajectories to prevent LLMs from engaging in redundant refinement rounds.
Specifically, we propose \textbf{\sys{}}, a multi-agent framework with dual-memory mechanisms. Its specialized agents collaborate through a closed-loop workflow for kernel generation and iterative refinement. In long-term memory, we store externalized expert optimization knowledge to guide LLMs in selecting concrete kernel optimization strategies, enabling traceable, audit-friendly decisions. In short-term memory, we maintain explicit optimization trajectories across iterations to help LLMs avoid repeated redundant optimization rounds.
This design enables interpretable method selection and robust, feedback-driven iterative improvement across heterogeneous kernels. 
Through extensive experiments on KernelBench \citep{ouyang2025kernelbenchllmswriteefficient} Levels 1--3, we demonstrate the effectiveness and robustness of \sys{}; it achieves a 100\% success rate and average speedups of 6.98$\times$, 7.89$\times$, and 1.94$\times$ over Torch Eager.

\section{Related Works}\label{sec:related_works}

Prior work has studied automated GPU kernel optimization through expert libraries, compiler/DSL ecosystems, and search-based autotuning, including cuDNN \citep{chetlur2014cudnnefficientprimitivesdeep}, TVM \citep{chen2018tvm}, Triton \citep{tillet2019triton}, and ThunderKittens \citep{spector2024thunderkittenssimplefastadorable}. These systems provide effective optimization primitives, but require substantial engineering effort and continuous maintenance as hardware and workloads evolve. Recent LLM-based methods reduce this burden by using code-generation models for kernel synthesis and optimization \citep{dong2025surveycodegenerationllmbased,Jiang_2026}. Training-based approaches, such as CUDA-L1, Kevin, and MTMC, adapt specialized models or policies using optimization signals \citep{li2025cudal1improvingcudaoptimization,baronio2025kevinmultiturnrlgenerating,zhu2025qimengkernelmacrothinkingmicrocodingparadigm}, but can be costly and may struggle to generalize across the large, hardware-dependent kernel optimization space \citep{zhai2024enabling}.

A complementary line performs inference-time closed-loop refinement, where LLMs iteratively revise kernels using compilation checks, correctness tests, runtime measurements, and profiling feedback \citep{wei2025astramultiagentgpukernel,lei2025pragmaprofilingreasonedmultiagentframework,zhang2025cudaforgeagentframeworkhardware,dong2025starkstrategicteamagents}. Recent methods further improve this process through expert-guided decomposition \citep{han2026eggexpertguidedagentframework}, failure-driven adaptation and diversity-preserving search \citep{du2026adaexplorefailuredrivenadaptationdiversitypreserving}, evolutionary program search \citep{sharma2025openevolve}, or DSL-constrained optimization with Speed-of-Light guidance \citep{hari2026improvingefficiencygpukernel}. However, these designs mainly structure the search process or accumulate optimization experience; they do not explicitly separate reusable cross-task knowledge from task-local trajectories through an auditable evidence-to-method interface. In contrast, \sys{} represents expert knowledge as deterministic decision rules retrieved jointly from static code features and profiling signals, while maintaining a separate short-term memory for repair and optimization histories. This functional separation enables traceable method selection and trajectory-aware refinement under a fixed optimization budget. A more detailed discussion is provided in Appendix~A of the supplementary material.

\section{Motivation}

\begin{figure*}[ht]
  \vskip 0.2in
  \begin{center}
    \centerline{\includegraphics[width=\textwidth, keepaspectratio]{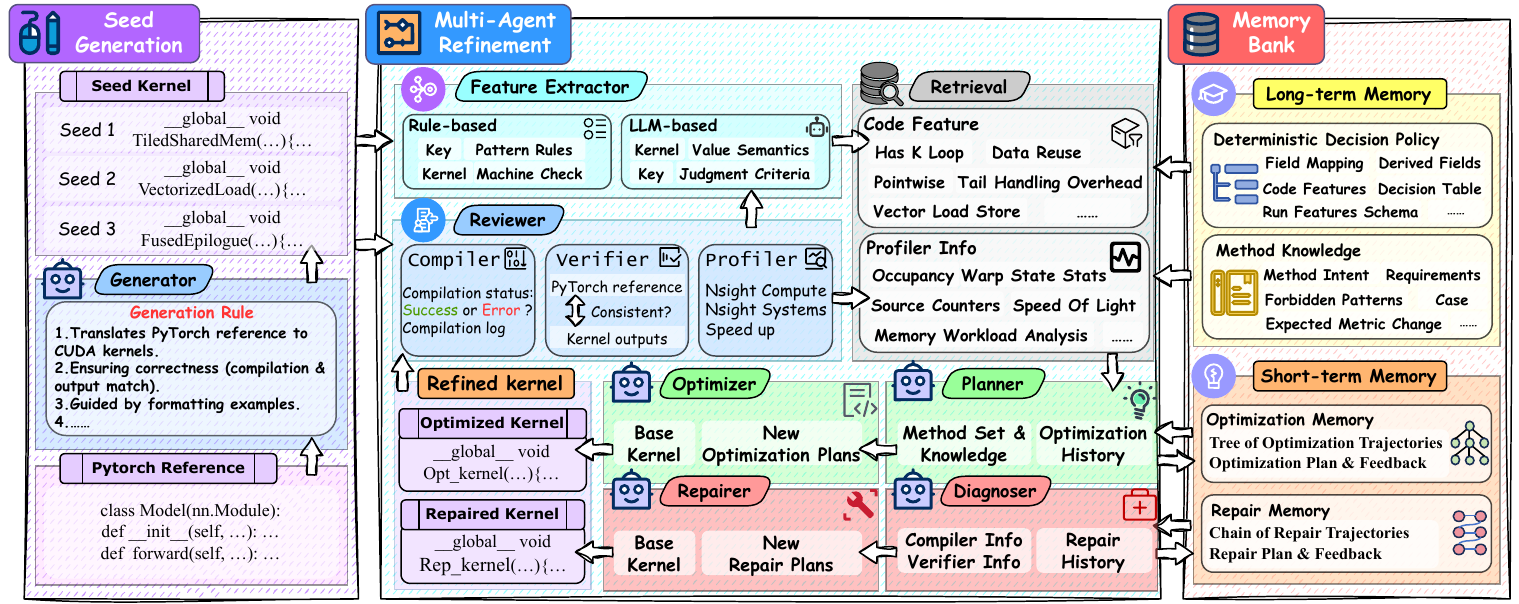}}
    \caption{
      Overview of \sys{}
    }
    \label{OverviewofKernelSkill}
  \end{center}
\end{figure*}
A concrete example helps illustrate why opaque optimization decisions and redundant refinement are problematic in LLM-based kernel optimization. 
Consider the KernelBench Level~2, Task~22 case \citep{ouyang2025kernelbenchllmswriteefficient}, whose implementation details are provided in Appendix~G the supplementary material.
This task applies a large $8192 \times 8192$ linear projection to a $1024 \times 8192$ input, followed by scaling, self-addition, clamping, a row-wise \texttt{logsumexp} reduction, and a final \texttt{mish}-based elementwise transformation. 
In this case, we observe that existing LLM-based kernel optimization methods often fail to identify the dominant performance bottleneck needed for selecting effective optimization strategies, and instead spend redundant refinement iterations repeating similar low-yield attempts. 
For example, a direct LLM-based optimization attempt is easily attracted to the visible fusion opportunity: it fuses \texttt{Linear} (GEMM), scaling, self-addition, and clamping into one CUDA kernel, while leaving \texttt{logsumexp} and the final \texttt{mish}-based transformation unfused. 
However, the fused kernel still implements GEMM as a naive global-memory dot-product loop without shared-memory tiling, vectorized loads, or tensor-core utilization, leaving the true bottleneck unresolved and achieving only \textbf{0.032$\times$} speedup. 
Later refinement rounds continue to explore similar low-yield fusion directions, showing that simply increasing the number of iterations does not resolve poor method selection or redundant refinement. 
By contrast, when the optimization is guided toward a cuBLASLt-based TF32 GEMM method and lightweight post-processing fusion, it resolves the main bottleneck and increases the speedup to \textbf{7.68$\times$}.
This motivates us to guide the optimization process toward effective optimization strategies by injecting explicit kernel optimization expertise, while avoiding redundant optimization rounds through recording and leveraging previous optimization trajectories.

\section{\sys{}}

We now present \sys{}, a memory-augmented multi-agent framework for GPU kernel optimization under profiling feedback. 
We first present an overview of \sys{}. We then elaborate on its multi-agent refinement pipeline, followed by a detailed illustration of the dual-memory mechanism.

\subsection{Overview}
An overview of \sys{} is shown in \cref{OverviewofKernelSkill}. Appendix~B of the supplementary material presents the pseudocode summarizing the complete execution loop of \sys{}.

\sys{} comprises a multi-agent pipeline and a dual-level memory system. The multi-agent pipeline contains Generator, Reviewer, Feature Extractor, Planner, Optimizer, Diagnoser and Repairer. 
The dual-level memory system includes long-term memory as the expert optimization knowledge base and short-term memory that preserves task-specific optimization and repair trajectories.

\sys{} first uses the Generator to produce a small set of seed kernels, which are then evaluated by the Reviewer; the best-performing seed is selected as the initial solution. Then, \sys{} starts a closed-loop refinement process.
At each round, the Reviewer evaluates the current kernel entering the
loop and returns structured feedback. The compilation status and
correctness results in the feedback determine the subsequent control
flow. If the kernel fails to compile or pass correctness verification,
the Diagnoser combines the observed failure signals with the
short-term repair memory to construct a repair plan, which the
Repairer applies to produce a revised kernel. If the kernel compiles
and passes correctness verification, \sys{} enters the optimization
branch. The Feature Extractor derives static code features, which are
combined with the profiling signals in the feedback to retrieve
applicable optimization methods from the long-term memory. The Planner
then integrates the retrieved methods with the short-term optimization
memory to formulate an optimization plan, which is executed by the
Optimizer.

The loop runs for at most $N$ rounds, and returns the best-performing kernel. We then introduce the key components of \sys{} in detail.

\subsection{Multi-Agent Pipeline}

\noindent\textbf{Generator.}
The Generator agent aims to generate functionally correct CUDA kernel implementations from a given PyTorch reference program while maximizing kernel diversity for subsequent optimization. Specifically, we implement the Generator as an LLM-based agent that translates PyTorch operators into equivalent custom CUDA kernels and prompts the LLM to explore diverse kernel implementations whenever possible.

\noindent\textbf{Reviewer.}
The Reviewer is a functional evaluation module that aims to assess each
generated CUDA kernel and return execution feedback for subsequent
repair and optimization. Given a generated kernel and its PyTorch
reference program, the feedback must capture both functional validity
and performance behavior, since the former determines whether the
kernel should be repaired, while the latter provides evidence for
identifying optimization opportunities. Accordingly, the Reviewer
reports the compilation status, correctness results, runtime
measurements, and profiling signals of the current implementation.
To produce this feedback, the Reviewer consists of three deterministic
components: a Compiler, a Verifier, and a Profiler. The Compiler builds
the generated kernel and reports whether compilation succeeds, together
with warnings and error messages when failures occur. The Verifier
checks functional correctness by comparing the kernel outputs with
those of the PyTorch reference under a prescribed numerical tolerance.
The Profiler measures execution latency and collects low-level runtime
signals using profiling tools such as \textbf{nsys} and \textbf{ncu},
including register usage, shared-memory footprint, and other
resource-related indicators. The resulting feedback is consumed by
downstream modules to determine the repair or optimization branch,
diagnose failures, identify performance bottlenecks, and guide
subsequent refinement.

\noindent\textbf{Feature Extractor.}
The Feature Extractor converts a CUDA kernel implementation into
source-level evidence for optimization-method retrieval. Profiling
signals reveal where performance bottlenecks occur, but they do not
always indicate whether a particular optimization method is
structurally applicable to the current implementation. To complement
this runtime evidence, the module extracts structural features of the code, such as memory-access patterns, precision choices and hardware-specific API/intrinsic usage.
These features are obtained directly from the kernel source without
executing the program. They are then combined with profiling signals
to form retrieval keys for selecting candidate optimization methods
from the long-term memory. For example, the applicability of Tensor
Core computation, vectorized memory access, shared-memory tiling, and
loop reorganization depends not only on the observed bottleneck but
also on source-level constraints that may be ambiguous from profiling
metrics alone.
To extract such evidence, the Feature Extractor combines rule-based
pattern matching for stable syntactic signatures with LLM-based
inference for semantically defined features whose surface forms vary
across implementations.


\noindent\textbf{Diagnoser.}
The Diagnoser is an LLM-based agent that aims to generate repair plans for invalid kernels. 
Given compiler feedback, verifier feedback, the current kernel, and the short-term repair memory, it outputs a structured plan that identifies likely root causes and suggests concrete fixes. 
It is triggered when a generated kernel fails compilation or violates output consistency, and the resulting plan is passed to the Repairer for code revision.
The Diagnoser interacts with short-term repair memory in a read--write manner. 
Before generating a repair plan, it reads previous repair attempts, including past failure signals, proposed fixes, revised kernels, and their outcomes. 
This history helps the Diagnoser avoid cyclic repair, where the agent repeatedly revisits known-failing edits. 
After the Repairer applies the current repair plan and the Reviewer evaluates the revised kernel, \sys{} writes the new repair attempt and its outcome back into short-term memory for subsequent rounds. 
We detail this mechanism in \cref{sec:dual_memory}.

\noindent\textbf{Planner.}
The Planner is an LLM-based agent that aims to select an optimization method and generate a concrete plan for a valid base kernel. 
After a kernel passes compilation and correctness verification, \sys{} retrieves candidate optimization methods from the long-term memory using its static code features and profiling feedback. 
Given the current base kernel, retrieved methods, method rationales, profiling feedback, and short-term optimization memory, the Planner outputs a stepwise optimization plan for the Optimizer.

The Planner reads short-term optimization memory before planning, including previously attempted methods, generated plans, revised kernels, and their performance outcomes. 
This history discourages repeated low-yield strategies and helps prioritize more promising optimizations. 
After the Optimizer applies the plan and the Reviewer evaluates the result, the new method attempt and its outcome are written back to short-term memory for subsequent rounds. 
We detail this mechanism in \cref{sec:dual_memory}.

\noindent\textbf{Optimizer and Repairer.}
The Optimizer and the Repairer are LLM-based code-editing agents that aim to execute upstream plans by producing revised CUDA kernels. 
Given the current kernel and a concrete plan, the Optimizer applies the optimization plan from the Planner to improve performance, while the Repairer applies the repair plan from the Diagnoser to fix compilation or correctness failures. 
Both agents are prompted to preserve the required kernel interface, code structure, formatting, and evaluation-specific constraints, so that the generated kernels can be directly compiled and executed in our pipeline. 
The revised kernel is then returned to the Reviewer for compilation, correctness checking, and profiling.

\subsection{Dual-Memory}\label{sec:dual_memory}
We present the detailed design of the long-term and short-term memory modules here.

\noindent\textbf{Long-term Memory.}
The long-term memory provides reusable expert knowledge for optimization-method selection across kernels. 
Given profiling feedback and static code features, \sys{} uses the long-term memory to retrieve feasible candidate methods and provide method-specific rationales and implementation cues for downstream planning and execution. 
This design allows \sys{} to select optimization methods through explicit evidence rather than relying solely on implicit prompt preferences or ad-hoc trial-and-error.

Concretely, the long-term memory consists of two complementary components:
\textbf{Deterministic Decision Policy}, which stores rule-based gating, prioritization, and method candidate selection criteria over profiling/code evidence; and
\textbf{Method Knowledge}, which stores method-specific rationales and implementation cues to make the selected actions interpretable and executable.
Given profiling feedback and code features, \sys{} first uses the Deterministic Decision Policy to filter and prioritize feasible method candidates, and then consults Method Knowledge to provide rationales and concrete implementation guidance for the Planner.
A full schema of the long-term memory fields is provided in Appendix~C of the supplementary material. We next describe how the two components support optimization-method retrieval.

\noindent\textbf{Optimization-method retrieval.}
\sys{} implements a structured retrieval pipeline that explicitly uses the two components of long-term memory. 
Given the current kernel's profiling signals and structural characteristics, including static code features, NCU metrics, and runtime features, \sys{} first normalizes raw profiling fields and derives composite indicators. 
These evidence fields are then matched against the Deterministic Decision Policy, where predicates, priority rules, and global veto rules are applied to filter and prioritize feasible optimization methods. 
After a candidate method set is selected, \sys{} consults Method Knowledge to attach method-specific rationales and implementation cues, which are then provided to the Planner for plan synthesis. 
During this process, method selection remains traceable: for each recommendation, \sys{} can record which normalized fields and predicates were satisfied, which rule or case in the decision table was matched, which constraints vetoed alternatives, and which method rationale was retrieved. 
The complete long-term memory decision workflow is provided in Appendix~D of the supplementary material. We next explain how this
knowledge is collected and organized into the two
long-term memory components.

\noindent\textbf{Expert knowledge sources.}
The long-term memory is built from expert kernel-optimization knowledge distilled by us. 
Concretely, we start from a large-scale GPU optimization survey \citep{hijma2023optimization} and translate its taxonomy into the two memory components through a three-step curation procedure.
(1) \textbf{Scenario abstraction:} we group optimization techniques into recurring, task-agnostic scenarios, such as memory-bound access patterns and underutilized parallelism, and identify the observable decision factors that practitioners use in bottleneck identification.
This step defines the high-level scenario and method taxonomy used by both Deterministic Decision Policy and Method Knowledge.
(2) \textbf{Evidence formalization:} we specify how each decision factor is measured from available signals, including normalized profiling metrics via \texttt{field\_mapping}, runtime features, and static code features; we also define deterministic composite indicators via \texttt{derived\_fields} so that the same concept can be evaluated consistently across kernels and tool versions.
This step mainly constructs the evidence fields used by the Deterministic Decision Policy.
(3) \textbf{Rule materialization:} we encode scenario-to-method criteria into predicates, priority rules, global veto rules, and a decision table that maps matched evidence patterns to candidate method sets, i.e., \texttt{allowed\_methods}.
These rules form the core of the Deterministic Decision Policy, while the companion \texttt{llm\_assist} store forms the Method Knowledge by providing method-level rationales and implementation cues.

This construction makes the knowledge reusable and auditable by expressing optimization expertise as task-independent scenarios, standardized evidence predicates, traceable evidence-to-rule matches, and method-specific implementation guidance. 
It separates deterministic method screening from semantic implementation support, improving selection stability while retaining LLM flexibility for plan synthesis.

\noindent\textbf{Short-term Memory.}
\begin{table*}[t]
  \caption{Success, Fast$_1$ and Speedup Results. \textbf{Success} is the percentage of tasks for which a method generates a kernel that compiles and passes correctness verification. \textbf{Fast$_1$} is the percentage of tasks whose generated kernel is at least as fast as the Torch baseline. \textbf{Speedup} is the average runtime speedup over Torch Eager. \textbf{\sys{}} achieves the best overall performance, matching the highest success rate and Fast$_1$ rate while delivering the largest speedup across all levels among all compared baselines.} 
  \label{Success_Speedup}
  \centering
  \small
  \begin{tabular}{lccccccccc}
    \toprule
    Method 
    & \multicolumn{3}{c}{Level 1} 
    & \multicolumn{3}{c}{Level 2} 
    & \multicolumn{3}{c}{Level 3} \\
    \cmidrule(lr){2-4} \cmidrule(lr){5-7} \cmidrule(lr){8-10}
    & Success$\uparrow$  &  Fast$_1$$\uparrow$ &  Speedup$\uparrow$ &  Success$\uparrow$ &  Fast$_1$$\uparrow$ &  Speedup$\uparrow$ &  Success$\uparrow$ &  Fast$_1$$\uparrow$ &  Speedup$\uparrow$ \\
    \cmidrule(lr){1-1} \cmidrule(lr){2-4} \cmidrule(lr){5-7} \cmidrule(lr){8-10}
    Torch Compile(default)& 1.00 & 0.61 & 1.51 & 1.00  & 0.76 & 1.22 & 1.00  & 0.68 & 1.40 \\
    Torch Compile(max-autotune)& 1.00 & 0.68 & 1.42 & 1.00  & 0.86 & 1.29 & 1.00  & 0.82 & 1.38 \\
    Kevin-32B & 0.69 & 0.04 & 0.84 & 0.67  & 0.08 & 0.40 & 0.46  & 0.02 & 0.32 \\
    AutoTriton & 0.36 & 0.09 & 1.20 & 0.55  & 0.22 & 0.96 & 0.56  & 0.26 & 0.83 \\
    Astra    & 0.95  & 0.43 & 1.48 & 0.98  & 0.73 & 0.99 & 0.93  & 0.35 & 0.90 \\
    PRAGMA    & 0.95  & 0.45 & 1.49 & 0.98 & 0.74 & 1.02 & 0.94  & 0.36 & 0.92 \\
    CudaForge & 0.96  & 0.54 & 1.45 & 1.00 & 0.89 & 2.10 & 0.96  & 0.68 & 1.28 \\
    QiMeng     & 1.00  & 0.59 & 2.20 & 0.99  & 0.66 & 1.22 & 0.70 & 0.40 & 0.73 \\
    OpenEvolve & 1.00  & 0.39 & 1.36 & 1.00 & 0.33 & 1.68 & 1.00  & 0.23 & 1.37 \\
    AdaExplore & 0.94  & 0.56 & 2.13 & 0.98  & 0.81 & 2.98 & 0.98 & 0.48 & 1.06 \\
    EGG & 1.00 & 0.70 & 1.79 & 1.00  & 1.00 & 2.80 & 1.00  & 0.86 & 1.48 \\
    Cursor & 1.00 & 0.34 & 1.61 & 1.00  & 0.51 & 2.74 & 1.00  & 0.19 & 1.36 \\
    STARK     & 1.00 & 0.71 & 3.03 & 1.00  & 1.00 & 2.69 & 1.00  & 0.87 & 1.58 \\
    \sys{}    & \textbf{1.00}  & \textbf{0.74}&\textbf{6.98} & \textbf{1.00}  & \textbf{1.00} & \textbf{7.89} & \textbf{1.00} & \textbf{0.88} & \textbf{1.94} \\
    \bottomrule
  \end{tabular}
\end{table*}
Short-term memory maintains an explicit per-task refinement trace, which enables consistent diagnosis and plan revision across rounds. It coordinates multi-agent iteration by conditioning the Diagnoser/Planner on past plans, edits, and outcomes. After each round, the short-term memory records: (i) the Planner's optimization plan and the Diagnoser's repair plan, (ii) the Compiler/Verifier's outcomes (compilation status and correctness check against the PyTorch reference), and (iii) the Profiler's performance feedback (e.g., latency and selected NCU/NSYS signals), together with the resulting kernel version. This trace is injected into the prompts of the Planner and the Diagnoser in subsequent rounds, so they can condition the next plan on what has already been tried and what feedback it produced.

Kernel optimization is a multi-round process: improvements typically emerge through successive, evidence-guided edits rather than a single modification. Without an explicit per-task trace, the Planner and Diagnoser may lose track of prior decisions and outcomes, leading to repeated ineffective attempts or reverting progress during later fixes. Short-term memory prevents this by maintaining a clear plan $\rightarrow$ result history, enabling more consistent plan revision across rounds and steadier refinement under profiling feedback.
We provide further details on its repair-stage and optimization-stage usage in Appendix~E of the supplementary material.

\section{Experiments}

\begin{table*}[t]
  \caption{Ablation Results. We ablate memory components: \textbf{Long-term memory} stores retrievable kernel optimization knowledge for reusable and explainable method selection, while \textbf{Short-term memory} records per-task optimization/repair trajectories to stabilize multi-round refinement. Removing either memory reduces Fast$_1$ and Speedup, confirming both components are critical to \textbf{\sys{}}'s improvements.}
  \label{Ablation}
  \centering
  \small
  \begin{tabular}{lccccccccc}
    \toprule
    Method 
    & \multicolumn{3}{c}{Level 1} 
    & \multicolumn{3}{c}{Level 2} 
    & \multicolumn{3}{c}{Level 3} \\
    \cmidrule(lr){2-4} \cmidrule(lr){5-7} \cmidrule(lr){8-10}
    & Success$\uparrow$ & Fast$_1$$\uparrow$ & Speedup$\uparrow$ & Success$\uparrow$ & Fast$_1$$\uparrow$ & Speedup$\uparrow$ & Success$\uparrow$ & Fast$_1$$\uparrow$ & Speedup$\uparrow$ \\
    \cmidrule(lr){1-1} \cmidrule(lr){2-4} \cmidrule(lr){5-7} \cmidrule(lr){8-10}
    w/o memory     & 0.96 & 0.46 & 1.52 & 0.98 &  0.75 & 1.05 & 0.94 &  0.36 & 0.95 \\
    w/o ST memory  & 0.96 & 0.51 & 3.53 & 0.98 & 0.85 & 2.42 & 0.94 & 0.70 & 1.78 \\
    w/o LT memory   & 1.00 & 0.55 &  1.77 & 1.00 & 1.00 & 1.18 & 1.00 & 0.79 & 1.04 \\
    \sys{}   & \textbf{1.00}  & \textbf{0.74}  & \textbf{6.98} & \textbf{1.00} & \textbf{1.00}  & \textbf{7.89} & \textbf{1.00} & \textbf{0.88}  & \textbf{1.94} \\
    \bottomrule
  \end{tabular}
\end{table*}

\subsection{Benchmark and Metrics}

KernelBench is an open-source benchmark for evaluating whether LLMs can generate \textit{correct} and \textit{efficient} GPU kernels for PyTorch workloads \citep{ouyang2025kernelbenchllmswriteefficient}. It organizes tasks into four difficulty levels: Level~1 contains single-kernel operators (100 tasks), Level~2 contains multi-operator workloads that require implementing and coordinating multiple operators/kernels (100 tasks), Level~3 includes full model architectures (50 tasks), and Level~4 targets end-to-end optimization of Hugging Face model architectures. Since prior LLM-based kernel generation and optimization methods are primarily evaluated on Levels~1--3, we follow prior work and report results on the 250 tasks from Levels~1--3.

KernelBench additionally provides metrics (e.g., Fast$_p$) that quantify the fraction of generated kernels that are both functionally correct and faster than a baseline by a specified threshold. In this work, we evaluate \sys{} and compare against prior approaches using the following metrics:
\ding{182} \textbf{Success:} the percentage of tasks for which the method generates a kernel that compiles and passes correctness verification.
\ding{183} \textbf{Fast$_1$:} the percentage of tasks for which the generated kernel is at least as fast as the Torch baseline.
\ding{184} \textbf{Speedup (vs. Torch Eager):} the average execution speedup across tasks, measured as the runtime ratio relative to Torch Eager (the unoptimized PyTorch implementation).


\subsection{Baselines}
\label{sec:baselines}

We compare \sys{} with representative compiler-based, training-based,
general coding-agent, and inference-time agentic approaches to GPU
kernel generation and optimization.

\noindent\textbf{Compiler and general coding baselines.}
We evaluate \texttt{torch.compile} under both its default and
max-autotune configurations as compiler-based optimization
baselines. We also include Cursor, a general-purpose coding agent that
can inspect source code, modify implementations, execute programs, and
iteratively respond to execution feedback
\citep{anysphere2026cursor}. These baselines assess whether general
compiler optimization or general-purpose agentic code editing is
sufficient for the KernelBench tasks without a kernel-specific
optimization workflow.

\noindent\textbf{Training-based methods.}
Kevin trains a specialized model using a multi-turn reinforcement
learning procedure that explicitly optimizes iterative kernel
refinement
\citep{baronio2025kevinmultiturnrlgenerating}. QiMeng adopts a
macro-to-micro paradigm that first produces high-level optimization
guidance and then realizes the selected strategy through stepwise
low-level code generation
\citep{zhu2025qimengkernelmacrothinkingmicrocodingparadigm}.

\noindent\textbf{Agentic and search-based optimization.}
CudaForge employs a training-free Coder--Judge workflow that
iteratively generates and evaluates CUDA kernels using correctness,
runtime, and hardware feedback
\citep{zhang2025cudaforgeagentframeworkhardware}. Astra organizes
kernel generation, testing, profiling, and planning through
specialized agents
\citep{wei2025astramultiagentgpukernel}, while PRAGMA explicitly maps
profiling evidence to bottleneck-specific optimization actions
\citep{lei2025pragmaprofilingreasonedmultiagentframework}. STARK
combines grounded optimization instructions, dynamic task context,
and strategic exploration for multi-round kernel refinement
\citep{dong2025starkstrategicteamagents}.
We additionally compare with more recent structured-search methods.
EGG decomposes kernel generation into algorithmic design and
hardware-specific refinement through expert-guided agent
collaboration
\citep{han2026eggexpertguidedagentframework}. AdaExplore combines
failure-driven adaptation with diversity-preserving tree search
\citep{du2026adaexplorefailuredrivenadaptationdiversitypreserving},
whereas OpenEvolve maintains an evolving population of programs and
generates new candidates from selected historical implementations
\citep{sharma2025openevolve}.

The main comparison includes both results reported in prior
publications and results obtained under our experimental setup.
Detailed result sources, reimplementation scope, and reproduction
settings are provided in Appendix~H of the supplementary material.

\subsection{Experimental Setting}
All primary KernelBench experiments reported in Table~1 are conducted on an NVIDIA A100 GPU with 80\,GB of memory. We use Python 3.11.13 and PyTorch 2.6.0+cu124 (CUDA 12.4). The CUDA runtime stack includes cuDNN 9.1.0 and cuBLAS 12.4. The experiments are implemented using the official KernelBench v0.1 codebase. Functional correctness is evaluated using the default numerical tolerance of KernelBench v0.1, with \texttt{atol}=\texttt{rtol}=$10^{-2}$.
For stable timing, we follow the evaluation procedure in prior work \citep{zhang2025cudaforgeagentframeworkhardware} by warming up with fixed input shapes before using CUDA events for timing; we use 25 warm-up iterations and 100 timed iterations and report the mean latency. Profiling signals are collected with NVIDIA Nsight Compute (ncu) and Nsight Systems (nsys).

\sys{} starts from seed kernels sampled by the Generator agent. For each task, we sample 3 candidate seeds and pick the best one (compiles, passes verification, and has the lowest mean latency) to start refinement. We run up to 15 refinement rounds per task, with temperature set to 1.0. All agent calls use ChatGPT-5.1 \citep{openai_chatgpt_gpt51_release_2025} as the base model. As detailed in Algorithm~1 in Appendix~B of the supplementary material, we promote a new base kernel only if it achieves at least 30\% relative speedup over the current base (rt=0.3) or an absolute speedup gain of at least 0.3 (at=0.3), to avoid unstable base updates from small fluctuations.

All locally reproduced and reimplemented LLM-based baselines use the
same GPT-5.1 backend and a maximum budget of 15 optimization rounds per
task. We preserve the method-specific agent organization, search
procedure, candidate context, and optimization mechanism whenever
applicable. Unless otherwise stated, the reproduced methods use the
same KernelBench version, correctness criteria, hardware environment,
and runtime-measurement procedure as \sys{}.

\subsection{Experimental Results and Analysis}

Table~\ref{Success_Speedup} reports the main results on KernelBench
Levels~1--3. In terms of Speedup, \sys{} achieves
6.98$\times$, 7.89$\times$, and 1.94$\times$ on the three levels,
respectively. The strongest baseline results are 3.03$\times$ from
STARK on Level~1, 2.98$\times$ from AdaExplore on Level~2, and
1.58$\times$ from STARK on Level~3. Accordingly, \sys{} improves over
the strongest baseline on each level by 130.4\%, 164.8\%, and 22.8\%,
respectively. These results indicate that evidence-grounded expert
method selection and trajectory-aware refinement enable \sys{} to
identify and realize higher-impact optimization strategies.

For Fast$_1$, \sys{} achieves 0.74, 1.00, and 0.88 on
Levels~1--3, respectively. It exceeds the strongest baselines on
Levels~1 and~3, improving over STARK from 0.71 to 0.74 and from
0.87 to 0.88, while tying EGG and STARK at 1.00 on Level~2. The
consistently strong Fast$_1$ results show that the acceleration of
\sys{} is distributed across a broad range of tasks rather than being
concentrated in only a few exceptionally fast kernels.

Finally, \sys{} achieves a Success rate of 1.00 on all three levels,
matching the highest correctness coverage among the compared methods.
This shows that \sys{} maintains strong reliability while delivering
superior optimization performance.

\subsection{Further Analysis}

To better understand the sources of the large performance gains
achieved by \sys{}, we manually inspect all generated kernels with a
Speedup greater than 10$\times$. For each case, we compare the kernel
implementation with the PyTorch reference and examine whether the
acceleration results from valid kernel-level optimization while
preserving the complete intended computation.

Most inspected high-speedup kernels are semantically valid. However,
we identify two anomalous kernels, from Levels~1 and~2, that pass the
official KernelBench correctness tests but do not preserve the intended
task semantics. Their apparent gains arise from benchmark-specific
shortcuts or omitted computation that is not exposed by the finite
test inputs and numerical tolerance. We do not observe analogous cases
among the other methods reproduced in our experiments. Nevertheless,
the underlying issue is not specific to \sys{}: these cases reveal
that passing the official KernelBench checker does not always guarantee
semantic equivalence to the PyTorch reference, and that sufficiently
strong search procedures may exploit such weaknesses in the evaluation
protocol.
We therefore discard the two anomalous kernels and, for each affected
task, select the fastest remaining semantically valid kernel from its
optimization trajectory. Table~\ref{tab:audited_kernelclaw} reports
the recomputed results.

\begin{table}[t]
  \caption{Speedup of \sys{} before and after excluding two
  anomalous kernels.}
  \label{tab:audited_kernelclaw}
  \centering
  \small
  \begin{tabular}{lccc}
    \toprule
    Result & Level 1 & Level 2 & Level 3 \\
    \midrule
    Original & 6.98$\times$ & 7.89$\times$ & 1.94$\times$ \\
    Audited  & 3.47$\times$ & 3.92$\times$ & 1.94$\times$ \\
    \bottomrule
  \end{tabular}
\end{table}

After excluding the two anomalous cases, \sys{} achieves Speedups of
3.47$\times$, 3.92$\times$, and 1.94$\times$ on Levels~1--3,
respectively. These values remain higher than the strongest baseline
results of 3.03$\times$, 2.98$\times$, and 1.58$\times$, corresponding
to relative improvements of 14.5\%, 31.5\%, and 22.8\%. Success and
Fast$_1$ remain unchanged because the replacement kernels from other
optimization rounds still compile, pass correctness verification, and
outperform Torch Eager. Thus, although the two anomalous kernels
substantially influence the original averages on Levels~1 and~2, the
overall advantage of \sys{} remains after auditing.

Complete task-level Speedup results after auditing are provided in Appendix~F. Token-consumption comparisons among the evaluated agentic methods are reported in Appendix~K. Detailed analyses of the two excluded cases and representative valid
high-speedup kernels are provided in Appendices~Q and R of the supplementary material.


\noindent\textbf{Generalization and robustness.} To further examine the robustness and generalizability of \sys{}, we
conduct five supplementary experiments. Specifically, we replace
GPT-5.1 with Qwen3-30B-A3B \citep{qwen3technicalreport} to study sensitivity to base-model
capability (Appendix~I), evaluate 30 sampled CUDABench \citep{zhu2026cudabenchbenchmarkingllmstexttocuda} tasks to examine
cross-benchmark transfer (Appendix~J), and evaluate 30 sampled TritonBench-G \citep{li-etal-2025-tritonbench} tasks to examine cross-backend applicability (Appendix~M).
We also rerun the complete optimization workflow on NVIDIA A40 and
RTX~4090 GPUs to evaluate hardware adaptation (Appendix~N), and compare \sys{} with
$\mu$CUTLASS+SOL \citep{hari2026improvingefficiencygpukernel} on the same 59 KernelBench tasks (Appendix~O). The results show that \sys{} requires
a sufficiently capable base LLM, while its optimization knowledge and
refinement workflow remain effective across benchmarks, programming
backends, and GPU platforms. It also outperforms the specialized
$\mu$CUTLASS+SOL baseline under the controlled setting. Detailed
experimental settings and results are provided in the supplementary material.

\subsection{Ablation}

Table~\ref{Ablation} evaluates three variants to isolate the effects of
the two memory components: (i) w/o memory, (ii) w/o long-term memory,
and (iii) w/o short-term memory. Removing both components degrades
correctness, optimization coverage, and acceleration across all three
levels. These results show that both memory mechanisms contribute to
improvements in Fast$_1$ and Speedup, while short-term memory also
improves Success.

Without short-term memory, some tasks still fail to produce a correct
kernel after 15 rounds, resulting in Success rates of 0.96, 0.98, and
0.94 on Levels~1--3. Enabling short-term memory raises Success to
1.00 on all three levels and improves Fast$_1$ from 0.51, 0.85, and
0.70 to 0.74, 1.00, and 0.88, respectively. It also improves Speedup
from 3.53$\times$, 2.42$\times$, and 1.78$\times$ to
6.98$\times$, 7.89$\times$, and 1.94$\times$. These gains are
consistent with its role in retaining previous failures, attempted
edits, and optimization outcomes, thereby allowing subsequent agents
to continue from the current trajectory rather than repeatedly
reconstructing task state. An additional experiment in Appendix~L of the supplementary material shows that short-term memory increases the proportion of optimization rounds under the same
15-round budget.

Long-term memory primarily determines the quality of the selected
optimization directions. Removing it preserves a Success rate of 1.00
but reduces Fast$_1$ to 0.55, 1.00, and 0.79 and, more substantially,
reduces Speedup from 6.98$\times$, 7.89$\times$, and 1.94$\times$
to 1.77$\times$, 1.18$\times$, and 1.04$\times$. Thus, the system
can still generate correct kernels without long-term memory, but it
rarely identifies transformations that yield large acceleration.
Together, these results show that long-term memory provides
high-impact expert guidance, while short-term memory enables that
guidance to be implemented and refined efficiently within a fixed
optimization budget.

\section{Conclusion}


We present \sys{}, a multi-agent framework for GPU kernel optimization
with dual-memory mechanisms. By combining reusable expert knowledge
with task-specific refinement trajectories, \sys{} supports
evidence-driven method selection and stable multi-round optimization.
Experiments on KernelBench demonstrate 100\% Success, the best or
tied-best Fast$_1$, and a Speedup advantage that remains after auditing. Limitations and broader deployment considerations
are discussed in Appendix~P of the supplementary material.

\appendix

\bibliography{aaai2027}


\section{Related Work}
\label{app:related_works}

\subsection{Automated GPU Kernel Optimization Prior to LLMs}

Before the emergence of LLM-based kernel generation, GPU performance
optimization was primarily supported by expert-maintained libraries,
compiler and domain-specific language infrastructures, and
search-based autotuning. Vendor libraries such as cuDNN provide
highly optimized implementations for a predefined collection of
operators \citep{chetlur2014cudnnefficientprimitivesdeep}. Compiler
and DSL systems, including TVM, Triton, and ThunderKittens, expose
hardware-aware scheduling abstractions and transformation primitives
for constructing efficient GPU programs
\citep{chen2018tvm,tillet2019triton,
spector2024thunderkittenssimplefastadorable}. Search-based autotuning
further explores combinations of schedules, tiling parameters, memory
layouts, and parallelization strategies using measurement feedback or
learned cost models \citep{chen2018learning}.

These systems provide essential abstractions for high-performance GPU
programming, but extending them to new operators, workload structures,
and hardware generations often requires substantial expert effort to
design transformation rules, scheduling spaces, and backend-specific
implementations. LLM-based approaches complement these systems by
directly generating or revising kernel code while retaining
compilation, correctness checking, and runtime measurement as
execution-grounded feedback.

Benchmarking efforts have also expanded beyond conventional CUDA
generation. TritonBench evaluates the correctness and efficiency of
LLM-generated Triton operators through both PyTorch-aligned tasks and
184 real-world operators collected from open-source repositories
\citep{li-etal-2025-tritonbench}. Its results highlight the continuing
difficulty of generating efficient GPU programs even when using a
higher-level programming abstraction such as Triton.

\subsection{Training-Based LLM Methods}

Recent progress in code-oriented LLMs has enabled specialized models
to synthesize and optimize GPU kernels through supervised adaptation,
reinforcement learning, and learned optimization policies
\citep{dong2025surveycodegenerationllmbased,Jiang_2026}.
CUDA-L1 and Kevin train models to improve CUDA generation and
multi-round kernel refinement
\citep{li2025cudal1improvingcudaoptimization,
baronio2025kevinmultiturnrlgenerating}. QiMeng introduces the
Macro-Thinking and Micro-Coding paradigm, in which high-level
optimization strategies are learned separately from their
step-by-step code realization
\citep{zhu2025qimengkernelmacrothinkingmicrocodingparadigm}.
Related approaches learn optimization behaviors at different levels
of the software stack, including Triton program synthesis and
low-level GPU instruction scheduling
\citep{li2025autotritonautomatictritonprogramming,
woo2025tritonrltrainingllmsthink,
he2025cuasmrloptimizinggpusass}.

Training-based methods can internalize recurring optimization
patterns and reduce reliance on manually designed inference-time
workflows. However, collecting suitable trajectories and performing
model training or reinforcement learning can be computationally
expensive. Moreover, optimization knowledge is primarily encoded in
model parameters or learned policies, making the relationship among
the observed runtime bottleneck, the selected optimization strategy,
and the resulting code transformation difficult to inspect. Their
learned policies may also require further adaptation when workloads,
programming abstractions, or target hardware change
\citep{zhai2024enabling}.

\subsection{Inference-Time Agentic and Search-Based Optimization}

A complementary line of work performs inference-time closed-loop
optimization. These systems iteratively generate and revise kernels
using compilation outcomes, correctness tests, runtime measurements,
and, when available, hardware profiling feedback. CudaForge adopts a
lightweight Coder--Judge workflow, while Astra organizes generation,
testing, profiling, and optimization planning through specialized
agents
\citep{zhang2025cudaforgeagentframeworkhardware,
wei2025astramultiagentgpukernel}. PRAGMA explicitly connects
profiling evidence with concrete optimization actions, and STARK
combines grounded optimization instructions, strategic exploration,
and dynamic task context
\citep{lei2025pragmaprofilingreasonedmultiagentframework,
dong2025starkstrategicteamagents}. Other studies extend closed-loop
optimization to Triton-oriented generation, evolutionary refinement,
and end-to-end PyTorch inference optimization
\citep{wang2025geakintroducingtritonkernel,
andrews2025gpukernelscientistllmdriven,
nagaitsev2025optimizingpytorchinferencellmbased}.

General-purpose coding agents can also serve as kernel-generation
baselines. Cursor, for example, provides repository search, source-code
editing, terminal execution, and iterative feedback within an
integrated coding environment \citep{anysphere2026cursor}. Such agents
offer flexible software-engineering capabilities but do not
intrinsically provide kernel-specific profiling interpretation,
hardware optimization rules, or structured mappings from runtime
evidence to applicable optimization methods.

Recent kernel optimizers introduce more structured mechanisms for
organizing inference-time exploration. EGG decomposes kernel
generation into algorithmic structure design and hardware-specific
tuning, using stage-aware multi-agent collaboration to progressively
refine the implementation
\citep{han2026eggexpertguidedagentframework}. AdaExplore combines
failure-driven adaptation with diversity-preserving search. It
converts recurring execution failures into reusable validity rules and
organizes candidate kernels as a tree that alternates between local
refinement and larger structural regeneration
\citep{du2026adaexplorefailuredrivenadaptationdiversitypreserving}.
OpenEvolve is a general evolutionary coding agent that maintains and
selects candidate programs and uses LLMs to generate new variations
under task-specific evaluation feedback
\citep{sharma2025openevolve}.

Another direction constrains or guides the search space using
domain-specific performance knowledge. $\mu$CUTLASS represents kernel
configurations, epilogue fusion, and multi-stage execution using a
compact CUTLASS-based DSL, while Speed-of-Light guidance estimates
performance headroom and allocates optimization effort accordingly
\citep{hari2026improvingefficiencygpukernel}. KernelEvolve formulates
kernel optimization as graph-based search and uses
retrieval-augmented prompt synthesis to adapt generation across
multiple programming abstractions and heterogeneous NVIDIA, AMD, and
custom accelerator platforms
\citep{liao2026kernelevolvescalingagentickernel}.

These methods demonstrate that staged decomposition, evolutionary
search, DSL constraints, and retrieval-augmented generation can
substantially improve kernel optimization. Their principal emphasis,
however, is on organizing candidate exploration, constraining the
implementation space, or improving the generation procedure.
In contrast, \sys{} focuses on making optimization-method selection
explicit and traceable: static source features and normalized runtime
evidence are jointly evaluated against deterministic applicability
rules before an optimization method is supplied to the code-generation
agents.

\subsection{Memory-Augmented Agents and Kernel Optimization}
\begin{algorithm*}[tb]
\caption{Multi-Agent Kernel Optimization with Dual-Level Memory}
\label{alg:kernel_opt}
\begin{algorithmic}
\STATE {\bfseries Input:} PyTorch reference $P$, Relative threshold $rt$, Absolute threshold $at$, max rounds $N$
\STATE {\bfseries Output:} Best optimized kernel $K_n$
\STATE $K_0 \leftarrow \textsc{Generator}(P)$
\STATE $(boolc_0,feedbackc_0) \leftarrow \textsc{Compiler}(K_0)$
\STATE $(boolv_0,feedbackv_0) \leftarrow \textsc{Verifier}(K_0)$
\STATE $(speedup_0,feedbackp_0) \leftarrow \textsc{Profiler}(K_0)$
\STATE Set $base\_kernel \leftarrow K_0$
\STATE Set $best\_kernel \leftarrow K_0$
\STATE Set $base \leftarrow 0$
\FOR{$i = 1$ {\bfseries to} $N$}
  \IF{not ($boolc_i$ and $boolv_i$)}
    \STATE $Repair\_plan \leftarrow \textsc{Diagnoser}(feedbackc_{i-1},feedbackv_{i-1},Repair\_memory)$
    \STATE $K_i \leftarrow \textsc{Repairer}(Repair\_plan,K_{i-1})$
  \ELSE
    \STATE $Code\_feature_{base} \leftarrow \textsc{FeatureExtractor}(base\_kernel)$
    \STATE $Methods \leftarrow \textsc{Retrieval}(Code\_feature_{base},feedbackp_{base})$
    \STATE $Optimization\_plan \leftarrow \textsc{Planner}(Methods,feedbackp_{base},Optimization\_memory)$
    \STATE $K_i \leftarrow \textsc{Optimizer}(Optimization\_plan,base\_kernel)$
  \STATE $(boolc_i,feedbackc_i) \leftarrow \textsc{Compiler}(K_i)$
  \STATE $(boolv_i,feedbackv_i) \leftarrow \textsc{Verifier}(K_i)$
  \STATE $(speedup_i,feedbackp_i) \leftarrow \textsc{Profiler}(K_i)$
  \STATE Update $Repair\_memory$
  \STATE Update $Optimization\_memory$
  \IF{$ speedup_i>speedup_{base}$}
    \STATE $best\_kernel \leftarrow K_i$
  \IF{$((speedup_i/speedup_{base}) > (1+rt)) or ((speedup_i-speedup_{base}) > at)$}
    \STATE $base\_kernel \leftarrow K_i$
    \STATE $base \leftarrow i$
      \ENDIF
    \ENDIF
  \ENDIF
\ENDFOR
\STATE {\bfseries Return} $best\_kernel$
\end{algorithmic}
\end{algorithm*}
Memory has become an important component of long-horizon LLM agents.
Recent surveys distinguish agent memory according to its
representation, function, and update dynamics, covering factual,
experiential, and working memory rather than relying exclusively on
the conventional long-term/short-term distinction
\citep{hu2026memoryageaiagents}. Agentic Memory, or AgeMem, integrates
long-term and short-term memory operations directly into the agent
policy. It allows the agent to learn when to store, retrieve, update,
summarize, and discard information through a reinforcement-learning
framework
\citep{yu-etal-2026-agentic}. These general-purpose approaches focus
primarily on long-horizon reasoning, autonomous context management,
and efficient use of finite context windows.

Memory and experience reuse have also been explored specifically for
accelerator and GPU kernel optimization. AccelOpt maintains an
optimization memory derived from previously observed slow--fast kernel
pairs and uses the accumulated experience to improve subsequent
kernels for emerging AI accelerators
\citep{zhang2026acceloptselfimprovingllmagentic}. KernelBlaster
constructs a persistent CUDA knowledge base from prior optimization
trajectories and uses memory-augmented in-context reinforcement
learning together with profile-guided textual feedback
\citep{dong2026kernelblastercontinualcrosstaskcuda}. AdaExplore
extracts reusable validity rules from recurring failures before
performing task-time candidate search
\citep{du2026adaexplorefailuredrivenadaptationdiversitypreserving}.
KernelEvolve retrieves relevant context to synthesize prompts for
kernel generation across operators, programming abstractions, and
hardware platforms
\citep{liao2026kernelevolvescalingagentickernel}. STARK additionally
maintains dynamic within-task context to support strategic refinement
across successive attempts
\citep{dong2025starkstrategicteamagents}.

Accordingly, the contribution of \sys{} is not simply the use of both
long-term and short-term memory. Instead, it defines a
kernel-specific functional separation between two forms of
optimization information. The long-term memory stores externally
curated expert knowledge as deterministic evidence predicates,
decision rules, and method-specific implementation guidance. Static
code features characterize whether a transformation is structurally
applicable, while profiling signals characterize the currently
observed runtime bottleneck. Their combination forms an explicit
evidence--rule--method chain for retrieving optimization knowledge.

The short-term memory serves a different role: it records the
task-local repair and performance-optimization trajectories, including
previous edits, selected methods, execution outcomes, and observed
failures. Repair and optimization histories are maintained separately
because they support different subsequent decisions. The former helps
avoid repeatedly producing invalid implementations, whereas the latter
helps the agents refine already correct kernels without revisiting
unproductive optimization attempts.

This design differs from AgeMem because \sys{} does not learn a
general policy for autonomous memory management. It also differs from
AccelOpt and KernelBlaster, whose reusable knowledge is primarily
accumulated from previously optimized kernels, and from AdaExplore,
whose cross-task memory focuses on validity rules distilled from
recurring failures. In \sys{}, the long-term memory is constructed
from explicit expert optimization knowledge and retrieved through
inspectable applicability conditions, while the short-term memory is
restricted to the current task. This functional decomposition supports
traceable method selection and trajectory-aware refinement without
requiring additional memory-policy training.

\section{Pseudocode of \sys{}}\label{app:pseudocode}

We provide the pseudocode of \sys{} in Algorithm~\ref{alg:kernel_opt}.

\section{Long-Term Memory Schema}\label{app:Longterm_schema}
\begin{figure*}[ht]
  \vskip 0.2in
  \begin{center}
    \centerline{\includegraphics[width=\textwidth, keepaspectratio]{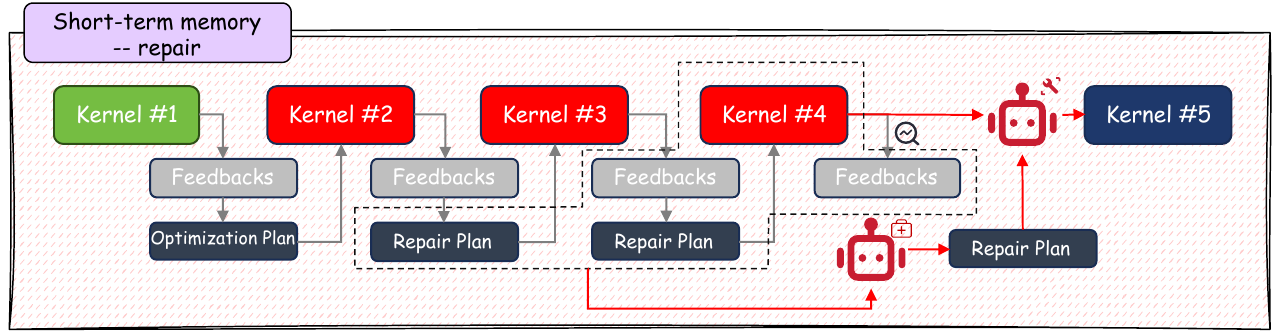}}
    \caption{
      The short-term memory for the current repair round
    }
    \label{Short-mem-repair}
  \end{center}
\end{figure*}

\begin{figure*}[ht]
  \vskip 0.2in
  \begin{center}
    \centerline{\includegraphics[width=\textwidth, keepaspectratio]{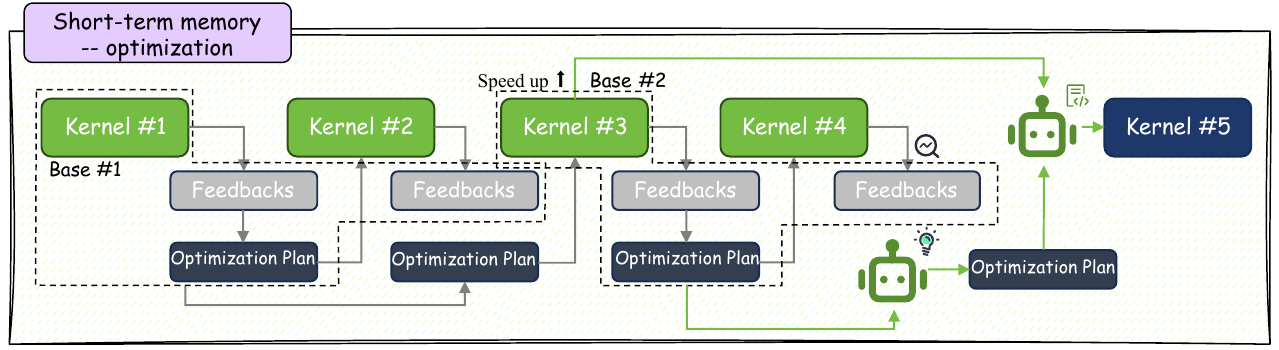}}
    \caption{
      The short-term memory for the current optimization round
    }
    \label{Short-mem-opt}
  \end{center}
\end{figure*}
The long-term memory mainly includes the following fields:

\ding{182} \textbf{field\_mapping}: maps raw NCU metrics to standardized fields for consistent downstream processing.
  
\ding{183} \textbf{run\_features\_schema}: defines runtime features extracted from Nsight Systems (e.g., \textbf{kernel\_launch\_count}).

\ding{184} \textbf{code\_features}: kernel-structure features extracted via deterministic scanning or LLM-assisted classification.

\ding{185} \textbf{derived\_fields}: deterministic fields derived from \textbf{field\_mapping}, \textbf{code\_features}, and \textbf{run\_features}.
  
\ding{186} \textbf{headroom\_tiers}: discretizes optimization headroom (e.g., High/Medium/Low) based on performance indicators.
  
\ding{187} \textbf{bottleneck\_priority\_rules}: resolves conflicts when multiple bottlenecks are detected.
  
\ding{188}\textbf{ncu\_predicates}: a library of reusable Boolean predicates over standardized NCU fields.
  
\ding{189}\textbf{global\_forbidden\_rules}: global veto rules that prevent unsafe or invalid optimizations.
  
\ding{190} \textbf{decision\_table}: maps bottleneck types, headroom tiers, and kernel characteristics to candidate optimization methods.
  
\ding{191} \textbf{llm\_assist} (Method Knowledge): Provides method-specific rationales and implementation cues for the candidate methods selected by the deterministic policy, enabling traceable decisions and faithful code-level execution.

\section{Long-Term Memory Decision Workflow}\label{app:Longterm_decision_workflow}
Specifically, the workflow is as follows:

\ding{182} \textbf{Input aggregation.}  Collect NCU metrics, runtime features, and code features for the current kernel.

\ding{183} \textbf{Metric normalization.} Map raw NCU metric keys to standardized field names via \textbf{field\_mapping}, making downstream decisions robust to tool-version-specific naming.

\ding{184} \textbf{Derived-field computation.}  
  Composite indicators are deterministically computed via \textbf{derived\_fields}, producing higher-level features that are more directly aligned with optimization decisions.

\ding{185} \textbf{Headroom tier assignment.}  
  The optimization headroom is categorized into discrete tiers (e.g., High/Medium/Low) using \textbf{headroom\_tiers}, which quantify the remaining optimization potential based on performance indicators.

\ding{186} \textbf{Bottleneck identification.}  
  The bottleneck type is identified by matching the kernel’s profiling signature against the \textbf{ncu\_signature} patterns defined in the \textbf{decision\_table}.
  
\ding{187} \textbf{Case matching.}  
  Given the bottleneck type and headroom tier, the system matches a specific decision case using conditions such as headroom tier, kernel structural properties, and additional gating predicates (\textbf{gate\_when}).
  
\ding{188}\textbf{Global rule enforcement.}  
  Before finalizing the recommendation, the system applies \textbf{global\_forbidden\_rules} as global veto constraints to prevent unsafe or invalid optimizations, regardless of local decision matches.

\ding{189} \textbf{Method set retrieval.}  
  The pipeline returns the final list of permitted optimization methods, i.e., the \textbf{allowed\_methods} associated with the matched decision case after global filtering.
  
\ding{190} \textbf{LLM-assisted planning.}  
  After deterministic gating, the system consults the \textbf{llm\_assist} knowledge base to provide non-binding explanations, implementation guidance, and expected benefits. This information helps the LLM interpret the selected methods and generate a concrete optimization plan.

\begin{figure*}[ht]
  \vskip 0.2in
  \begin{center}
    \centerline{\includegraphics[width=\textwidth, keepaspectratio]{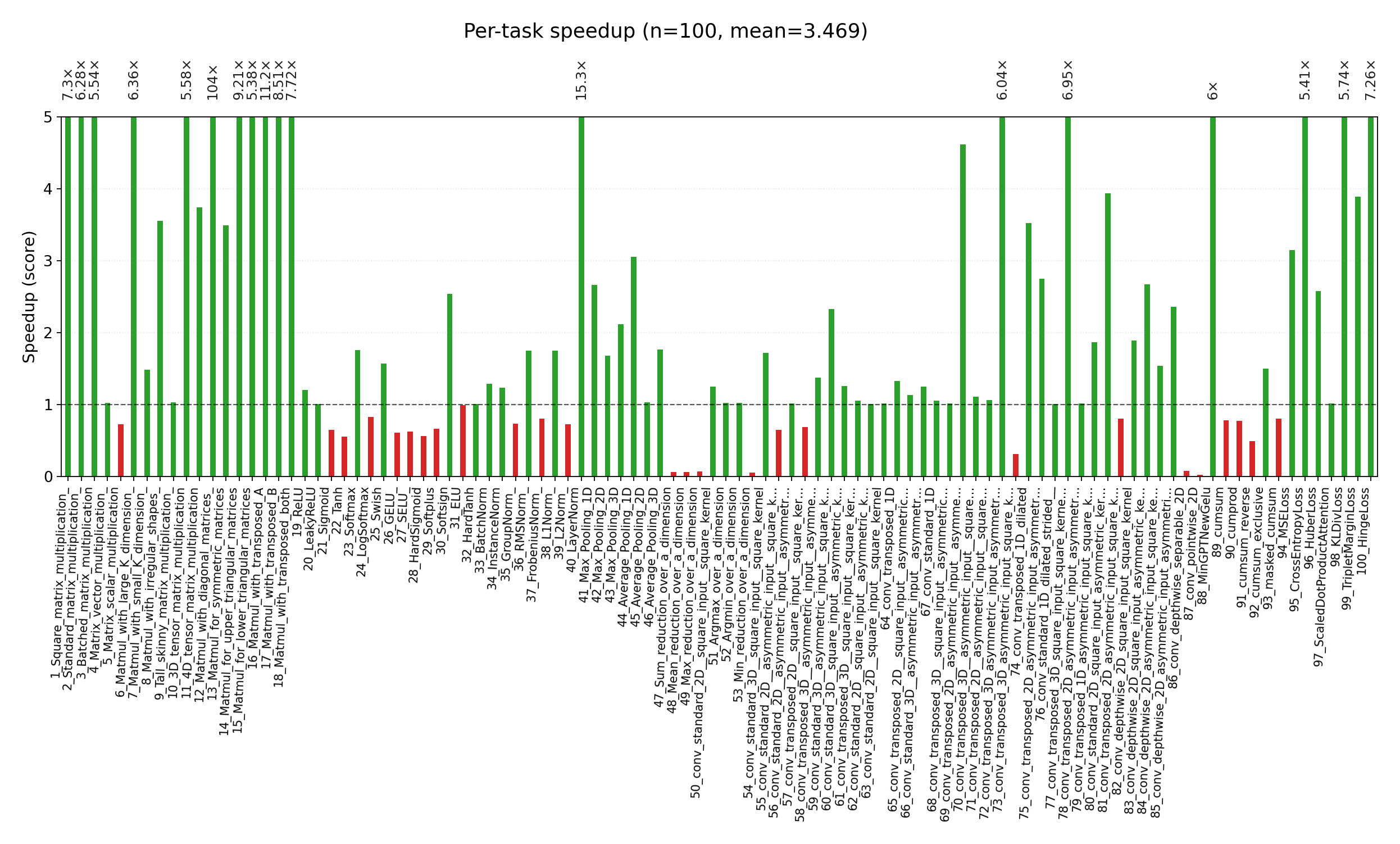}}
    \caption{
      Speedup distribution on 100 tasks at Level~1.
    }
    \label{fig:level1-speedup-distribution}
  \end{center}
\end{figure*}
\section{Short-Term Memory in Repair and Optimization}\label{app:short-term-memory}

This appendix details how short-term memory is organized and used in the repair and optimization branches. 
Although both branches maintain a per-task refinement trace, they differ in how the base kernel is selected and how the memory is updated.

\paragraph{Repair Stage.}
During the repair stage, the Diagnoser utilizes short-term memory to generate repair plans. At this stage, the short-term memory may consist of multiple chained memory segments. Each chain begins with a kernel that first fails compilation or correctness verification. As illustrated in \cref{Short-mem-repair}, the chain starts from kernel \#2, and in each subsequent iteration, the most recently generated kernel is used as the base kernel for the next repair step.

Importantly, although each repair iteration operates on the latest kernel version, the repair plan itself is generated based on the entire history of repair attempts and their corresponding outcomes within the chain. For example, kernel \#5 is generated by the Repairer based on a repair plan produced by the Diagnoser, which in turn is derived from the accumulated feedback spanning from kernel \#2 to kernel \#4. Thus, the repair process is guided jointly by historical repair knowledge and the most recent execution feedback, enabling more informed and stable error correction.

\paragraph{Optimization Stage.}
The use of short-term memory in the optimization stage differs from that in the repair stage, particularly in how the base kernel is selected. During repair, the base kernel is always the most recent kernel in the chain. In contrast, during optimization, the base kernel must first pass compilation and correctness verification.
Whether the base kernel should be updated is determined by two hyperparameters: a relative speedup threshold and an absolute speedup threshold. The relative speedup measures the percentage improvement of the newly generated kernel over the current base kernel, while the absolute speedup measures the absolute difference in execution performance. If either threshold is exceeded, the base kernel is updated accordingly.

As illustrated in \cref{Short-mem-opt}, all optimization methods previously applied to the same base kernel, along with their execution results, are stored in the short-term memory and incorporated into the context for generating the next optimization plan. Based on this plan, the Optimizer produces a new kernel by iterating on the current base kernel, which is then evaluated in the next optimization round.

\section{Task-Level Speedup Distributions Across Levels}\label{app:speedup_distributions}
This appendix reports task-level speedup distributions across the three evaluation levels. Each figure corresponds to one level and presents the speedup values achieved on 100 individual tasks within that level. These distributions complement the aggregate results in the main paper by showing how optimization gains vary across tasks, rather than only reporting averaged performance.
\clearpage
\onecolumn
\begin{figure*}[ht]
  \vskip 0.2in
  \begin{center}
    \centerline{\includegraphics[width=\textwidth, keepaspectratio]{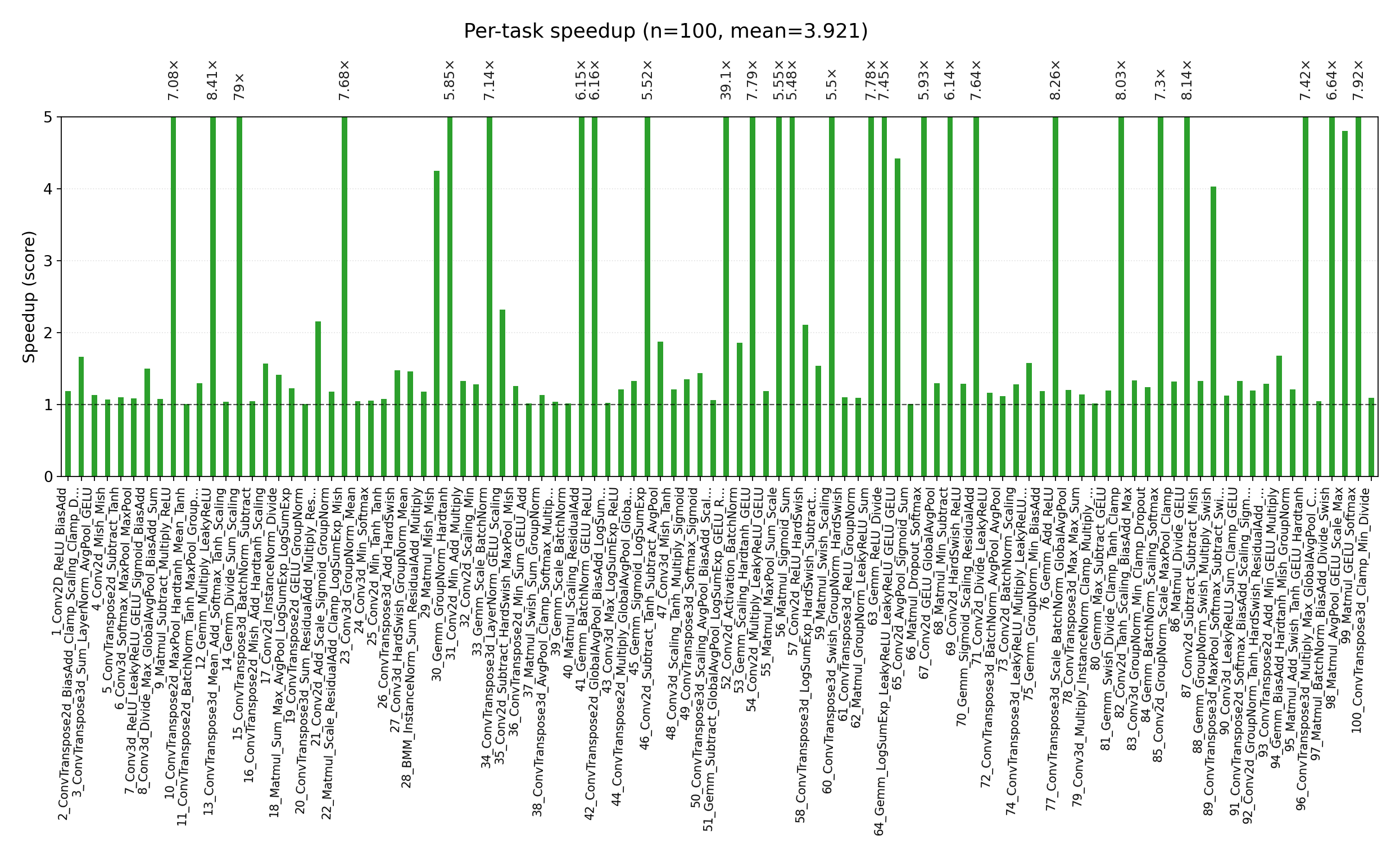}}
    \caption{
      Speedup distribution on 100 tasks at Level~2.
    }
    \label{fig:level2-speedup-distribution}
  \end{center}
\end{figure*}

\begin{figure*}[ht]
  \vskip 0.2in
  \begin{center}
    \centerline{\includegraphics[width=\textwidth, keepaspectratio]{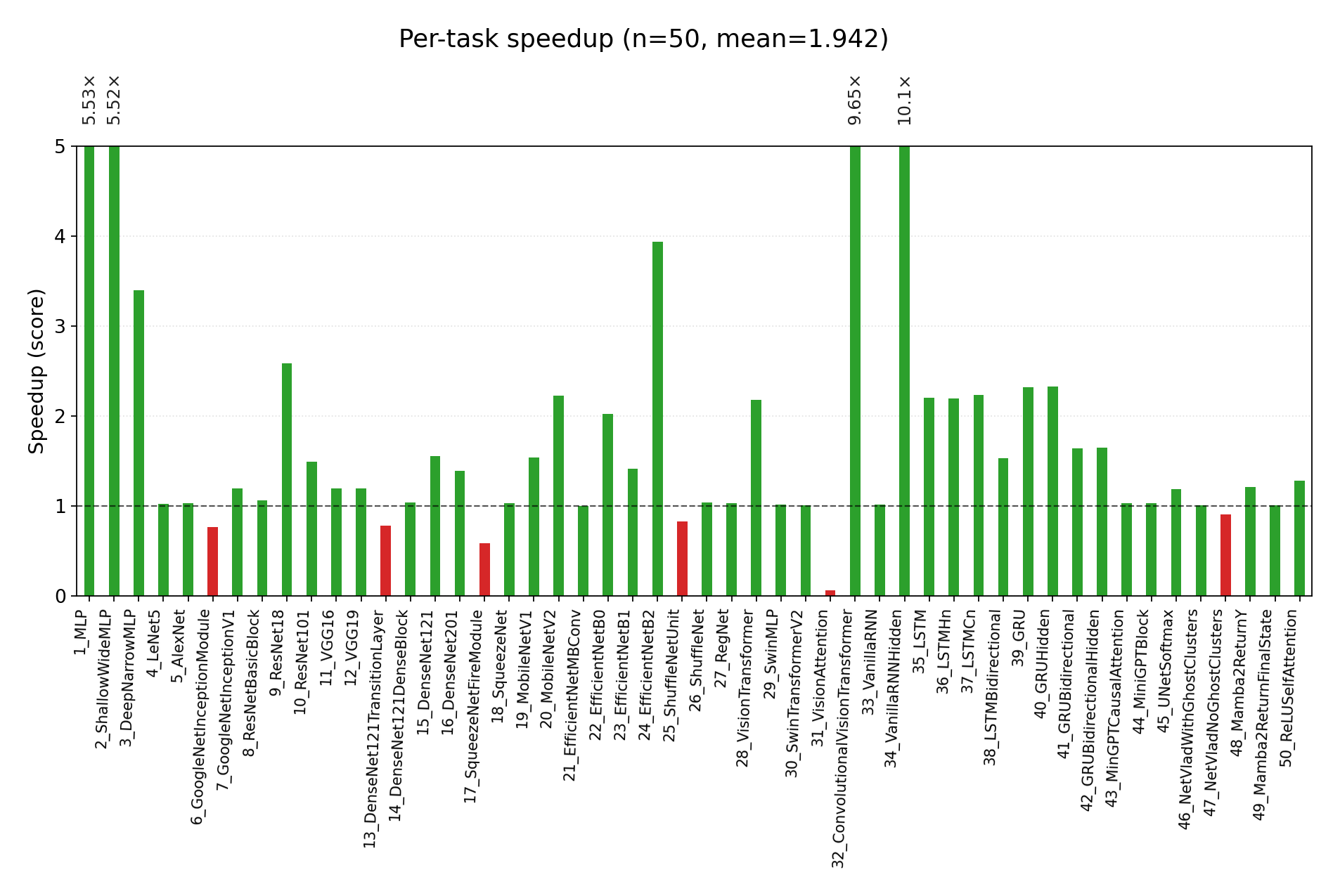}}
    \caption{
      Speedup distribution on 50 tasks at Level~3.
    }
    \label{fig:level3-speedup-distribution}
  \end{center}
\end{figure*}

\section{Motivating KernelBench Case}\label{app:motivating_case}
\begin{center}
\captionof{listing}{Task code for the motivating KernelBench case.}
\end{center}

\begin{Verbatim}[fontsize=\footnotesize,breaklines=true,breakanywhere=true]
import torch
import torch.nn as nn

class Model(nn.Module):
    """
    Model that performs a matrix multiplication, scales the result, 
    adds a residual connection, clamps the output,
    applies LogSumExp, and finally applies the Mish activation function.
    """
    def __init__(self, input_size, hidden_size, scale_factor, clamp_min, clamp_max):
        super(Model, self).__init__()
        self.matmul = nn.Linear(input_size, hidden_size)
        self.scale_factor = scale_factor
        self.clamp_min = clamp_min
        self.clamp_max = clamp_max

    def forward(self, x):
        """
        Args:
            x: Input tensor of shape (batch_size, input_size).

        Returns:
            Output tensor of shape (batch_size, hidden_size).
        """
        x = self.matmul(x)
        x = x * self.scale_factor
        x = x + x
        x = torch.clamp(x, self.clamp_min, self.clamp_max)
        x = torch.logsumexp(x, dim=1, keepdim=True)
        x = x * torch.nn.functional.mish(x)  # Mish activation
        return x
\end{Verbatim}

\begin{center}
\captionof{listing}{Optimized kernel generated without \sys{} for the motivating KernelBench case.}
\end{center}

\begin{Verbatim}[fontsize=\footnotesize,breaklines=true,breakanywhere=true]
__global__ void fused_matmul_scale_clamp_kernel(
    const float* __restrict__ X,
    const float* __restrict__ W,
    const float* __restrict__ B,
    float* __restrict__ Out,
    int batch,
    int in_features,
    int out_features,
    float scale_factor,
    float clamp_min,
    float clamp_max
){
    int row = blockIdx.x * blockDim.x + threadIdx.x; 
    int col = blockIdx.y * blockDim.y + threadIdx.y; 

    if(row < batch && col < out_features){
        float acc = 0.0f;

        for(int k = 0; k < in_features; k++){
            acc += X[row * in_features + k] * W[col * in_features + k];
        }
        acc += B[col];

        acc = acc * scale_factor * 2.0f;
        acc = fminf(fmaxf(acc, clamp_min), clamp_max);

        Out[row * out_features + col] = acc;
    }
}
torch::Tensor fused_matmul_scale_clamp(
    torch::Tensor X,
    torch::Tensor W,
    torch::Tensor B,
    float scale_factor,
    float clamp_min,
    float clamp_max
){
    const int batch = X.size(0);
    const int in_features = X.size(1);
    const int out_features = W.size(0);

    auto Out = torch::empty({batch, out_features}, X.options());

    dim3 block(16,16);
    dim3 grid((batch + block.x -1)/block.x,
              (out_features + block.y -1)/block.y);

    fused_matmul_scale_clamp_kernel<<<grid, block>>>(
        X.data_ptr<float>(),
        W.data_ptr<float>(),
        B.data_ptr<float>(),
        Out.data_ptr<float>(),
        batch,
        in_features,
        out_features,
        scale_factor,
        clamp_min,
        clamp_max
    );

    return Out;
}
\end{Verbatim}

\twocolumn

\section{Result Sources and Reproducibility}
\label{app:result_sources}

\paragraph{Result sources.}
The baseline results are obtained either from the original
publications or from experiments conducted under our setup. Since
complete executable implementations of STARK and QiMeng were not
available, we report their best KernelBench results from the
corresponding papers
\citep{dong2025starkstrategicteamagents,
zhu2025qimengkernelmacrothinkingmicrocodingparadigm}. These results
are included as published reference points and were not obtained under
our unified experimental environment.

All other baseline results are produced under our experimental setup.
For CudaForge, AdaExplore, and OpenEvolve, we adapt their released
implementations to our model interface and evaluation pipeline
\citep{zhang2025cudaforgeagentframeworkhardware,
du2026adaexplorefailuredrivenadaptationdiversitypreserving,sharma2025openevolve}. Since complete executable
implementations of Astra, PRAGMA, EGG, and $\mu$CUTLASS+SOL were not
available, we reimplement these methods according to the architectures,
algorithms, prompts, and experimental descriptions provided in their
papers
\citep{wei2025astramultiagentgpukernel,
lei2025pragmaprofilingreasonedmultiagentframework,
han2026eggexpertguidedagentframework,
hari2026improvingefficiencygpukernel}. We preserve their core agent workflows,
search procedures, and optimization mechanisms while adapting the
input, model, and evaluation interfaces to our KernelBench setting.

\paragraph{Reproduction settings.}
All reproduced and reimplemented LLM-based methods use GPT-5.1 and
the same maximum budget of 15 optimization rounds per task. They are
evaluated using the same KernelBench version, correctness criteria,
numerical tolerances, runtime-measurement procedure, and hardware
environment. Method-specific agent organizations, search strategies,
and candidate-context mechanisms are retained whenever applicable.

Unless otherwise stated, all ablation and supplementary results are
also obtained under our experimental setup. These include the
short-term-memory and token-cost analyses, the controlled comparison
with $\mu$CUTLASS+SOL, the cross-benchmark evaluation on CUDABench,
the cross-backend evaluation on TritonBench-G, and the cross-GPU
evaluation on the NVIDIA A40 and RTX~4090.

\section{Effect of Base LLM Capability}\label{app:effect_base_llm}

This appendix analyzes the sensitivity of our multi-agent optimization framework to the capability of the underlying LLM. As shown in Table~\ref{qwen30b}, when replacing the base model with Qwen3-30B-A3B and using 15-round refinement budget, the overall performance drops substantially across all three levels. The degradation is mainly caused by a low rate of generating kernels that both compile and pass correctness verification. This failure occurs not only during the initial seed-kernel generation stage, but also during subsequent refinement rounds, where the model is given an optimization plan but still frequently fails to produce a valid optimized kernel. Consequently, both the Success rate and the average Speedup remain low.

These results suggest that our framework requires the underlying LLM to possess a minimum level of CUDA code generation ability, optimization reasoning, and instruction-following reliability. We also tested Llama-3.1-70B and Qwen3-0.6B under the same prompting protocol. In addition to compilation errors and correctness failures, these models sometimes produced empty outputs or non-code natural-language responses instead of the required code block, indicating weaker adherence to the output-format constraints. For example, Llama-3.1-70B even produced an unrelated aphoristic sentence, "It is not the strongest of the species that survives, nor the most intelligent that survives. It is the one that is most adaptable to change." -- Charles Darwin, rather than a CUDA kernel. Therefore, the effectiveness of the proposed multi-agent framework depends not only on the agent workflow itself, but also on whether the base LLM can reliably follow structured prompts and generate syntactically valid, semantically faithful CUDA kernels.

\begin{table}[t]
  \caption{Results with Qwen3-30B-A3B under 15-round optimization budget.}
  \label{qwen30b}
  \centering
  \small
  \begin{tabular}{lccc}
    \toprule
    KernelBench
    & \multicolumn{1}{c}{Success$\uparrow$} 
    & \multicolumn{1}{c}{Fast$_1$$\uparrow$} 
    & \multicolumn{1}{c}{Speedup$\uparrow$} \\
    \cmidrule(lr){1-1}  \cmidrule(lr){2-2} \cmidrule(lr){3-3} \cmidrule(lr){4-4}
    Level 1 & 0.51 &  0.10 &  0.60 \\
    Level 2   & 0.07 &  0.00 &  0.01 \\
    Level 3    & 0.04 &  0.00  &  0.03  \\
    \bottomrule
  \end{tabular}
\end{table}

\section{Cross-Benchmark Evaluation on CUDABench}\label{app:cross_bench_eva}

This appendix evaluates whether the proposed multi-agent optimization framework and its kernel-optimization knowledge can generalize beyond the KernelBench task set. To this end, we conduct an additional experiment on CUDABench \cite{zhu2026cudabenchbenchmarkingllmstexttocuda}. Unlike KernelBench, CUDABench defines each task through a natural-language description and requires the system to generate a functionally correct CUDA kernel from that description. Therefore, the granularity and specificity of the natural-language prompt can strongly affect whether a valid kernel can be generated. Since improving natural-language-to-kernel generation accuracy, such as pass@1 or pass@3, is not the primary objective of \sys{}, we use this experiment mainly to examine whether our optimization framework and retrieved expert knowledge remain useful for diverse operator-level optimization tasks outside KernelBench.

Specifically, we focus on Level~1 tasks in CUDABench, where the natural-language descriptions are relatively detailed, and randomly sample 30 tasks for evaluation. For each task, \sys{} is given a 10-round budget, including both seed-kernel generation and subsequent refinement. We report two types of performance measurements: the original CUDABench-Score defined by CUDABench, and the speedup metric adapted from KernelBench, where runtime improvement is measured against the corresponding \texttt{bench.cu} implementation provided by CUDABench.

Table~\ref{tab:cudabench-sample-statistics} summarizes the overall statistics on the 30 sampled tasks. \sys{} successfully generates kernels that pass consistency checking for 21 tasks, which then enter the iterative optimization stage. Among these valid tasks, 17 are matched with at least one retrieved expert optimization method, indicating that the collected expert knowledge has broad coverage over the sampled CUDABench operators. Within these 17 tasks, 14 achieve an improvement in CUDABench-Score, and 12 improve both CUDABench-Score and speedup. The remaining 3 tasks receive retrieved optimization methods but fail to generate a correct optimized kernel within the 10-round budget after seed generation. These results suggest that, when \sys{} retrieves a suitable optimization method and the underlying LLM can generate a semantically correct kernel, the framework can effectively improve CUDABench-Score and often improve both evaluation dimensions simultaneously.

We further ablate the retrieved expert knowledge by disabling method retrieval and relying only on short-term memory for iterative refinement. Under this setting, only 7 of the same 17 tasks achieve CUDABench-Score improvements, as shown in Table~\ref{tab:cudabench-sample-statistics}. This comparison indicates that the retrieved expert methods provide optimization guidance beyond what the LLM can achieve using only local search history and its internal knowledge of kernel optimization.

Finally, Table~\ref{tab:cudabench-representative-cases} provides representative CUDABench cases from the 30 sampled tasks where matching appropriate optimization methods leads to substantial performance improvements. These cases show that expert-method retrieval can produce large gains across different operator patterns, such as FIR filtering, AdamW, matrix multiplication, and Gaussian blur. The comparison with the no-retrieval setting further illustrates that these gains are not merely due to local iterative refinement, but are closely associated with retrieving suitable optimization knowledge for the target operator.

\begin{table}[t]
\centering
\small
\caption{Statistics on 30 randomly sampled CUDABench tasks.}
\label{tab:cudabench-sample-statistics}
\begin{tabular}{lcc}
\toprule
\textbf{Stage / Outcome} & \textbf{Count} & \textbf{Rate} \\
\midrule
Sampled tasks & 30 & -- \\
Valid seed kernels & 21 / 30 & 70.0\% \\
Matched expert methods & 17 / 21 & 81.0\% \\
Improved CUDABench-Score & 14 / 17 & 82.4\% \\
Improved both Score and speedup & 12 / 17 & 70.6\% \\
Failed after matched methods & 3 / 17 & 17.6\% \\
\midrule
w/o expert methods: improved Score & 7 / 17 & 41.2\% \\
\bottomrule
\end{tabular}
\end{table}

\begin{table*}[t]
\centering
\small
\caption{Representative CUDABench cases improved by matched optimization methods. 
The last column reports the CUDABench-Score improvement obtained when expert-method retrieval is disabled.}
\label{tab:cudabench-representative-cases}
\begin{tabular}{c p{4.5cm} c c c c}
\toprule
\textbf{Task} 
& \textbf{Matched Method} 
& \textbf{Seed $\rightarrow$ Best} 
& \textbf{$\Delta$} 
& \textbf{Relative} 
& \textbf{w/o Expert Methods $\Delta$} \\
\midrule
\texttt{358} 
& \texttt{CMP\_Irregular\_Compute\_S2}; \texttt{ML\_REUSE} 
& 0.174 $\rightarrow$ 0.727 
& +0.553 
& +318\% 
& +0.097 \\

\texttt{254} 
& \texttt{MBW\_*}; \texttt{ML\_FALLBACK} 
& 0.521 $\rightarrow$ 0.916 
& +0.395 
& +76\% 
& +0.171 \\

\texttt{119} 
& \texttt{GEMM}; \texttt{CUBLASLT}; \texttt{TILING} 
& 0.211 $\rightarrow$ 0.437 
& +0.226 
& +107\% 
& +0.000 \\

\texttt{79} 
& \texttt{CMP\_Irregular\_Compute\_S2} 
& 0.150 $\rightarrow$ 0.351 
& +0.201 
& +134\% 
& +0.061 \\
\bottomrule
\end{tabular}
\end{table*}

\section{Token Cost}
\label{app:token_cost}

We compare the token consumption of the reproduced agentic kernel
optimization methods under a unified experimental setting. All methods
use GPT-5.1 as the underlying language model and are evaluated for
15 iterative optimization rounds per task. Each round generates and
evaluates one new candidate kernel. For OpenEvolve, we retain its
default evolutionary context, which includes up to three
top-performing programs and two diverse programs when generating a
new child. For AdaExplore, we retain its default candidate-context
settings, with a pool size of five for large-step reconstruction and
up to five recent ancestor kernels for small-step refinement.

Token usage is collected from the API usage records and accumulated
over all LLM calls during each optimization run. We separately report
input tokens, output tokens, and their sum. The values in
Table~\ref{tab:token_cost} denote the average token consumption per
task and are reported in thousands of tokens.

\begin{table}[t]
  \caption{Average token consumption per task using GPT-5.1 under a
  unified budget of 15 optimization rounds. All values are reported
  in thousands of tokens.}
  \label{tab:token_cost}
  \centering
  \small
  \begin{tabular}{lrrr}
    \toprule
    Method & Input $\downarrow$ & Output $\downarrow$ & Total $\downarrow$ \\
    \midrule
    CudaForge  & 51k & 20k & 71k \\
    Astra      & 56k & 21k & 77k \\
    PRAGMA     & 53k & 21k & 74k \\
    AdaExplore & 84k &  9k & 93k \\
    OpenEvolve & 58k & 12k & 70k \\
    EGG        & 55k & 22k & 77k \\
    \sys{}     & 68k & 23k & 91k \\
    \bottomrule
  \end{tabular}
\end{table}

Table~\ref{tab:token_cost} shows that the compared methods remain
within a similar overall range of language-model consumption under
the same 15-round budget. AdaExplore incurs the highest input-token
cost, as its tree-search prompts incorporate a candidate pool or
recent ancestral trajectory when proposing each new kernel.
OpenEvolve maintains a lower total cost despite conditioning each
generation on multiple historical programs. \sys{} consumes more
input tokens than several closed-loop baselines because its prompts
include both retrieved long-term optimization knowledge and
task-specific repair and optimization histories. Nevertheless, its
total token consumption remains comparable to the search-based
methods, indicating that its performance improvements do not arise
from a larger number of candidate evaluations.

\section{Effect of Short-Term Memory on Optimization-Round Utilization}
\label{app:stm_round_utilization}

The short-term memory records task-specific repair and optimization
trajectories, allowing the agents to condition subsequent decisions on
previous attempts and their outcomes. To examine whether this design
improves the allocation of the fixed optimization budget, we measure
the proportion of rounds devoted to performance optimization rather
than compilation or correctness repair.

Specifically, a round is categorized as a \emph{repair round} when it
attempts to resolve a compilation failure, runtime error, or incorrect
output. A round is categorized as an \emph{optimization round} when it
starts from a functionally correct kernel and aims to further reduce
its execution latency. We report the optimization-round ratio, defined
as the number of optimization rounds divided by the total number of
executed rounds. All variants use the same maximum budget of 15 rounds
per task.

\begin{table}[t]
  \caption{Percentage of rounds devoted to performance optimization
  under the same 15-round budget. Higher values indicate that a larger
  fraction of the trajectory is used to improve already correct
  kernels rather than repair invalid ones.}
  \label{tab:stm_round_utilization}
  \centering
  \small
  \begin{tabular}{lcccc}
    \toprule
    Method
    & L1
    & L2
    & L3
    & Avg. \\
    \midrule
    w/o short-term memory & 67.9\% & 74.7\% & 70.4\% & 71.0\% \\
    \sys{}                & 73.2\% & 78.2\% & 72.1\% & 74.5\% \\
    \bottomrule
  \end{tabular}
\end{table}

As shown in Table~\ref{tab:stm_round_utilization}, short-term memory
increases the optimization-round ratio across all three KernelBench
levels. The improvements are 5.3, 3.5, and 1.7 percentage points on
Levels 1, 2, and 3, respectively, yielding an average increase of
3.5 percentage points. Since all variants are constrained to the same
15-round budget, the higher ratio indicates that short-term memory
helps the system resolve invalid intermediate kernels more efficiently
and allocate a larger portion of the remaining trajectory to
performance optimization. This improved budget utilization provides
one explanation for the stronger final speedups achieved by the full
system.

\section{Cross-Backend Evaluation on TritonBench-G}
\label{app:tritonbench}

To evaluate whether \sys{} generalizes beyond the CUDA backend used
in the main experiments, we further evaluate it on TritonBench-G,
which contains real-world Triton operators collected from open-source
repositories. This experiment examines whether the expert knowledge
retrieval and trajectory-based refinement mechanisms of \sys{} remain
effective when applied to Triton implementations.
\begin{table}[t]
\centering
\small
\caption{Optimization outcomes on 30 randomly sampled
TritonBench-G tasks.}
\label{tab:tritonbench-sample-statistics}
\begin{tabular}{lcc}
\toprule
Outcome & Count & Rate \\
\midrule
Sampled tasks
& 30
& -- \\

Valid seed kernels
& 30 / 30
& 100.0\% \\

Tasks with matched expert methods
& 29 / 30
& 96.7\% \\

Tasks improved by \sys{}
& 23 / 29
& 79.3\% \\

Tasks not improved by \sys{}
& 6 / 29
& 20.7\% \\

\midrule
Tasks improved w/o expert methods
& 11 / 29
& 37.9\% \\
\bottomrule
\end{tabular}
\end{table}

As shown in Table~\ref{tab:tritonbench-sample-statistics},
\sys{} successfully matches at least one applicable expert method
for 29 of the 30 sampled Triton operators. Among these matched tasks,
23 obtain a faster kernel than their original Triton implementation,
corresponding to an improvement rate of 79.3\%. This result indicates
that the source-level features and expert optimization knowledge used
by \sys{} remain applicable when the target implementation is written
in Triton rather than CUDA.

We additionally disable expert-method retrieval while keeping the
underlying model, optimization workflow, and 15-round budget
unchanged. Under this ablation, only 11 of the 29 matched tasks are
improved. The improvement rate therefore decreases from 79.3\% to
37.9\%, showing that the retrieved expert methods provide useful
guidance for Triton kernel refinement rather than merely adding
descriptive context to the prompts.

\begin{table*}[t]
\centering
\small
\caption{Representative TritonBench-G tasks improved by the expert
methods retrieved by \sys{}. Speedup is measured relative to the
original Triton seed kernel.}
\label{tab:tritonbench-representative-cases}
\begin{tabular}{c p{3.4cm} c c c c}
\toprule
\textbf{Task}
& \textbf{Matched Expert Methods}
& \textbf{w/o Expert Methods}
& \textbf{\sys{}}
& \textbf{Absolute Gain}
& \textbf{Relative Gain} \\
\midrule

\texttt{geglu\_tanh\_triton}
& \texttt{Tune\_Num\_Warps};
  \texttt{Constexpr\_Block\_Tuning}
& 4.71$\times$
& 7.33$\times$
& +2.62$\times$
& +55.6\% \\

\texttt{batched\_vecmat\_mult}
& \texttt{Rewrite\_Mulsum\_to\_tl\_dot};
  \texttt{Enable\_tl\_dot\_tf32}
& 1.25$\times$
& 6.07$\times$
& +4.82$\times$
& +385.6\% \\

\texttt{dequantize\_matmul}
& \texttt{Quant\_Dequant\_Matmul};
  \texttt{Tune\_Num\_Warps}
& 0.99$\times$
& 4.55$\times$
& +3.56$\times$
& +359.6\% \\

\texttt{layernorm\_fwd\_triton}
& \texttt{tl\_rsqrt};
  \texttt{Welford\_Layernorm}
& 1.28$\times$
& 1.80$\times$
& +0.52$\times$
& +40.6\% \\

\midrule
Mean over 30 tasks
& --
& 1.38$\times$
& 1.90$\times$
& +0.52$\times$
& +37.7\% \\
\bottomrule
\end{tabular}
\end{table*}

Table~\ref{tab:tritonbench-representative-cases} provides several
representative examples. Across all 30 sampled tasks, \sys{} achieves an average speedup of 1.90$\times$, compared with 1.38$\times$ when expert-method retrieval
is disabled. These results demonstrate that \sys{} can transfer its
evidence-driven method retrieval and trajectory-based refinement
mechanisms to real-world Triton operators, extending its applicability
beyond the CUDA kernels evaluated in the main experiments.

\section{Cross-GPU Evaluation}
\label{app:cross_gpu}

To evaluate the hardware adaptability of \sys{}, we conduct additional
experiments on two GPU platforms, NVIDIA A40 and NVIDIA GeForce RTX
4090. For each GPU, \sys{} independently performs the complete
15-round kernel-generation and optimization procedure using the target
GPU's runtime and profiling feedback. Generated kernels are compiled
for the native architecture of the corresponding GPU.

We follow the same KernelBench correctness and runtime-measurement
protocol as in the main experiments. In particular, Success measures
the fraction of tasks that compile and pass the official correctness
tests, Fast$_1$ measures the fraction of correct kernels that
outperform Torch Eager, and Speedup is computed relative to Torch
Eager measured on the same GPU. Therefore, the reported speedups
reflect hardware-specific improvements over each GPU's own baseline
rather than absolute runtime differences between GPUs.

\begin{table}[t]
  \caption{Cross-GPU evaluation of \sys{} on KernelBench. Speedup is
  measured relative to Torch Eager on the corresponding GPU.}
  \label{tab:cross_gpu}
  \centering
  \small
  \begin{tabular}{llccc}
    \toprule
    GPU & Level & Success $\uparrow$ & Fast$_1$ $\uparrow$
        & Speedup $\uparrow$ \\
    \midrule
    \multirow{3}{*}{A40}
      & Level 1 & 1.00 & 0.68 & 5.86$\times$ \\
      & Level 2 & 1.00 & 0.98 & 6.14$\times$ \\
      & Level 3 & 1.00 & 0.70 & 1.81$\times$ \\
    \midrule
    \multirow{3}{*}{RTX 4090}
      & Level 1 & 1.00 & 0.71 & 5.48$\times$ \\
      & Level 2 & 1.00 & 0.94 & 5.60$\times$ \\
      & Level 3 & 1.00 & 0.68 & 1.71$\times$ \\
    \bottomrule
  \end{tabular}
\end{table}

As shown in Table~\ref{tab:cross_gpu}, \sys{} achieves a Success rate
of 1.00 on all three KernelBench levels on both GPUs, demonstrating
that the generated kernels remain functionally correct across
different hardware platforms. It also maintains high Fast$_1$ values,
particularly on Level~2, where 98\% and 94\% of the tasks outperform
the corresponding Torch Eager baselines on A40 and RTX~4090,
respectively.

The performance trends are also consistent across the two GPUs.
\sys{} achieves substantial speedups on Levels~1 and~2, reaching
5.86$\times$ and 6.14$\times$ on A40 and 5.48$\times$ and
5.60$\times$ on RTX~4090. The gains on Level~3 are smaller but remain
stable, with speedups of 1.81$\times$ and 1.71$\times$, respectively.
These results indicate that \sys{} does not rely on optimization
patterns specific to the A100 platform used in the main experiments.
Instead, its profiling-guided method retrieval and iterative refinement
procedure can adapt kernel implementations to the performance
characteristics of different target GPUs.

\section{Comparison with $\mu$CUTLASS+SOL}
\label{app:mucutlass_sol}
We further compare \sys{} with $\mu$CUTLASS+SOL, a recent approach
that combines a compact CUTLASS-based domain-specific language with
Speed-of-Light guidance for efficient kernel optimization
\citep{hari2026improvingefficiencygpukernel}. We reproduce $\mu$CUTLASS+SOL on
the same 59 KernelBench tasks used in its original evaluation. For a
controlled comparison, both methods use GPT-5.1 and are given the same
budget of 15 candidate-generation and evaluation rounds per task. We
apply the same correctness-checking and runtime-measurement protocol
to both methods and report Success and geometric-mean speedup over
Torch Eager.

\begin{table}[t]
  \caption{Controlled comparison with $\mu$CUTLASS+SOL on the same
  59 KernelBench tasks under GPT-5.1 and a 15-round optimization
  budget.}
  \label{tab:mucutlass_sol}
  \centering
  \small
  \begin{tabular}{lcc}
    \toprule
    Method
    & Success $\uparrow$
    & Geo. Mean Speedup $\uparrow$ \\
    \midrule
    $\mu$CUTLASS+SOL & 1.00 & 1.99$\times$ \\
    \sys{}           & 1.00 & 2.21$\times$ \\
    \bottomrule
  \end{tabular}
\end{table}

As shown in Table~\ref{tab:mucutlass_sol}, both methods successfully
generate correct kernels for all 59 tasks. Under the same 15-round
budget, \sys{} achieves a geometric-mean speedup of 2.21$\times$,
compared with 1.99$\times$ for $\mu$CUTLASS+SOL, corresponding to an
11.1\% relative improvement. This result demonstrates that the
evidence-driven expert-method retrieval and iterative refinement of
\sys{} remain effective when compared with a specialized
DSL- and SOL-guided optimization approach.

\section{Limitations}
\label{app:Limitations}
This work has several limitations. First, the effectiveness of
\sys{} depends on the coverage of its curated expert knowledge; unseen
optimization opportunities may not match the current method taxonomy
or decision rules. Second, the framework still relies on the underlying
LLM to correctly implement retrieved optimization methods, and weaker
models may produce invalid or suboptimal kernels.
Moreover, finite correctness tests cannot fully exclude numerical errors, race conditions, unsafe memory accesses, or benchmark-specific shortcuts. We perform an
additional semantic audit during evaluation, but production deployment
would still require stronger task-specific verification.

Future work will expand the expert knowledge base and integrate stronger
compiler-assisted verification and repair.

\section{Audited Case Analysis for Excluded Kernels}
\label{app:audited_case}

This appendix presents representative anomalous extreme-speedup cases that are excluded from the audited results reported with $^\dagger$. These cases achieve apparently large speedups but do not constitute valid CUDA kernel optimizations because their outputs pass the benchmark checker through non-equivalent computation or degenerate task structure. 

Specifically, the Level~1, Task~100 (Algorithm~\ref{alg:case_hacking1}) case reveals a statistical masking problem rather than a correct semantic optimization. The reference computation broadcasts the target vector over the prediction matrix and averages the hinge loss over all $32768 \times 32768$ elements. However, the generated kernel flattens both tensors and sets the reduction length to \texttt{targets.numel()}, thereby computing the loss only over the first row of the prediction matrix. This is mathematically non-equivalent to the reference. Nevertheless, it can pass the benchmark consistency check because the prediction matrix is sampled i.i.d. from a uniform distribution and the row length is very large. As a result, the mean loss over a single row is numerically close to the mean loss over the full matrix under the tested random distribution, allowing the semantic error to remain within the checker tolerance.

The Level~2, Task~80 (Algorithm~\ref{alg:case_hacking2}) case should not be considered a genuine optimized kernel. The generated implementation does not accelerate the intended GEMM-reduction-activation pipeline; instead, it exploits a degenerate configuration of the reference program. Since the max operation reduces the feature dimension to length one with \texttt{keepdim=True}, the subsequent mean over the same dimension is identical to the tensor itself, making the subtraction always produce zero. As a result, GELU also returns zero, and the final output becomes independent of both the input and the linear-layer parameters. A kernel that directly returns zeros can therefore pass final-output consistency checks, but its speedup comes from bypassing the computation rather than optimizing it. This case should be treated as a degenerate shortcut or benchmark artifact, not as evidence of valid CUDA kernel optimization. 

These cases illustrate why we audit extreme-speedup results and exclude kernels whose measured gains arise from semantic shortcuts, checker artifacts, or degenerate benchmark configurations rather than genuine kernel-level optimization.

\section{Audited High-Speedup Kernel Optimization Cases}
\label{app:audited_high_speedup_case}

This appendix presents representative high-speedup kernels generated by \sys{} that pass our auditing procedure. Unlike the anomalous cases excluded in Appendix~\cref{app:audited_case}, these examples preserve the intended task semantics while achieving substantial acceleration through valid algebraic simplification and kernel fusion. 

In the first case, the original computation performs a large matrix multiplication followed by division, summation over the output dimension, and scaling. The optimized kernel observes that the final summation allows the GEMM to be algebraically collapsed: instead of materializing the full intermediate matrix, it first reduces the weight matrix into a column-sum vector and then computes a row-wise dot product between the input and this vector, while fusing the division and scaling into a single scalar factor. This avoids constructing a large intermediate activation and significantly reduces memory traffic and arithmetic work. 

In the second case, the optimized kernel similarly exploits the downstream reduction structure. Since the global average over the GEMM output depends only on the summed weights, summed bias, and the mean of the subtraction parameter, the full GEMM--subtract--pool pipeline can be rewritten as a row-wise dot product followed by scalar normalization, GELU, and broadcast residual addition. The generated CUDA kernel fuses these operations into one execution path, avoiding multiple framework-level kernels and intermediate tensors. 

Together, these cases illustrate that the audited high-speedup results are driven by genuine kernel-level optimization, including algebraic reduction, operation fusion, reduced intermediate materialization, and more efficient parallel execution.

\section{Artifact Use and Implementation Details}
\label{app:artifacts-implementation}

\paragraph{Artifacts, licenses, and implementation.}
We use KernelBench, CUDABench, and TritonBench-G solely for research
evaluation. KernelBench is released under the MIT License, and we cite
its creators following the official citation instructions. For
CUDABench, TritonBench-G, PyTorch, Triton, the CUDA Toolkit, NVIDIA
Nsight Systems and Nsight Compute, and the released implementations of
the applicable baselines, we comply with their respective licenses,
terms of use, and citation requirements.

The benchmark tasks are used only for their intended research and
evaluation purposes. We do not redistribute third-party benchmark
assets, source code, or model artifacts except where explicitly
permitted by their licenses. Our released artifacts are limited to our
implementation, baseline reproductions, configuration files,
evaluation scripts, prompts, generated kernels, and derived
experimental records. Access to commercial LLM backends is obtained
through their official APIs, and no proprietary model weights are
redistributed.

The artifacts used in this work consist of program code,
kernel-optimization tasks, compiler and profiler outputs, and
aggregated execution measurements. They contain no human-subject data
or personally identifying information. Our implementation uses
PyTorch, CUDA, Triton, \texttt{nsys}, and \texttt{ncu}, together with
the evaluated LLM backends. Detailed software versions, hardware
configurations, optimization budgets, result provenance, and
evaluation protocols are provided in the experimental setup and the
supplementary material.

\clearpage
\onecolumn

\begin{center}
\captionof{listing}{Excluded anomalous extreme-speedup case for Level~1, Task~100.}
\label{alg:case_hacking1}
\end{center}
\begin{Verbatim}[fontsize=\footnotesize,breaklines=true,breakanywhere=true]
import torch
import torch.nn as nn
from torch.utils.cpp_extension import load_inline

source = r"""
#include <torch/extension.h>
#include <cuda_runtime.h>
#include <ATen/cuda/CUDAContext.h>
#include <c10/cuda/CUDAException.h>

__global__ void hinge_loss_reduce_kernel(
    const float* __restrict__ preds,
    const float* __restrict__ targets,
    float* __restrict__ out,
    int64_t N
){
    float thread_sum = 0.0f;
    for (int64_t idx = blockIdx.x * blockDim.x + threadIdx.x;
         idx < N;
         idx += (int64_t)blockDim.x * gridDim.x)
    {
        float v = 1.0f - preds[idx] * targets[idx];
        if (v > 0.0f) thread_sum += v;
    }
    extern __shared__ float sdata[];
    sdata[threadIdx.x] = thread_sum;
    __syncthreads();

    for (int s = blockDim.x / 2; s > 0; s >>= 1){
        if (threadIdx.x < s){
            sdata[threadIdx.x] += sdata[threadIdx.x + s];
        }
        __syncthreads();
    }
    if (threadIdx.x == 0){
        atomicAdd(out, sdata[0]);
    }
}
torch::Tensor hinge_loss_forward(torch::Tensor preds, torch::Tensor targets){
    TORCH_CHECK(preds.dtype() == torch::kFloat32, "preds must be float32");
    TORCH_CHECK(targets.dtype() == torch::kFloat32, "targets must be float32");
    TORCH_CHECK(preds.is_contiguous(), "preds must be contiguous");
    TORCH_CHECK(targets.is_contiguous(), "targets must be contiguous");

    auto N = targets.numel();
    TORCH_CHECK(preds.numel() >= N, "preds must have at least as many elements as targets");

    auto out = torch::zeros({1}, preds.options());

    const int block_size = 256;
    int grid_size = (N + block_size - 1) / block_size;
    grid_size = max(1, min(grid_size, 4096));

    size_t shmem = block_size * sizeof(float);
    auto stream = at::cuda::getDefaultCUDAStream();

    hinge_loss_reduce_kernel<<<grid_size, block_size, shmem, stream>>>(
        preds.data_ptr<float>(),
        targets.data_ptr<float>(),
        out.data_ptr<float>(),
        N
    );
    C10_CUDA_KERNEL_LAUNCH_CHECK();
    return out[0] / (float)N;
}
"""

cpp_src = "torch::Tensor hinge_loss_forward(torch::Tensor preds, torch::Tensor targets);"

hinge_loss_mod = load_inline(
    name="hinge_loss_cuda_fused",
    cpp_sources=cpp_src,
    cuda_sources=source,
    functions=["hinge_loss_forward"],
    verbose=True,
    extra_cflags=["-O3", "-std=c++17"],
    extra_cuda_cflags=[
        "-gencode=arch=compute_80,code=sm_80",
        "--expt-relaxed-constexpr",
        "-lineinfo",
    ],
    extra_ldflags=[]
)

class ModelNew(nn.Module):
    def __init__(self):
        super().__init__()
        self.holder = nn.Module()  
        self.kernel = hinge_loss_mod

    def forward(self, predictions, targets):
        if predictions.dim() > 1:
            predictions = predictions.reshape(-1)
        if targets.dim() > 1:
            targets = targets.reshape(-1)

        if not predictions.is_contiguous():
            predictions = predictions.contiguous()
        if not targets.is_contiguous():
            targets = targets.contiguous()

        return self.kernel.hinge_loss_forward(predictions, targets)

\end{Verbatim}

\begin{center}
\captionof{listing}{Excluded anomalous extreme-speedup case for Level~2, Task~80.}
\label{alg:case_hacking2}
\end{center}
\begin{Verbatim}[fontsize=\footnotesize,breaklines=true,breakanywhere=true]
import torch
import torch.nn as nn
from torch.utils.cpp_extension import load_inline

cuda_src = r"""
#include <torch/extension.h>
#include <cuda_runtime.h>
#include <ATen/cuda/CUDAContext.h>
#include <c10/cuda/CUDAException.h>

__global__ void zero_fill_kernel(float* __restrict__ out, int64_t n) {
    int64_t i = (int64_t)blockIdx.x * blockDim.x + threadIdx.x;
    if (i < n) out[i] = 0.0f;
}

torch::Tensor zero_output_like_batch(torch::Tensor x) {
    auto out = torch::empty({x.size(0), 1}, x.options());
    int64_t n = out.numel();
    int threads = 256;
    int blocks = (int)((n + threads - 1) / threads);
    auto s = at::cuda::getDefaultCUDAStream();
    zero_fill_kernel<<<blocks, threads, 0, s>>>(out.data_ptr<float>(), n);
    C10_CUDA_KERNEL_LAUNCH_CHECK();
    return out;
}
"""

cpp_src = "torch::Tensor zero_output_like_batch(torch::Tensor x);"

mod = load_inline(
    name="k80_zero_ext",
    cpp_sources=cpp_src,
    cuda_sources=cuda_src,
    functions=["zero_output_like_batch"],
    extra_cflags=["-O3", "-std=c++17"],
    extra_cuda_cflags=["-O3", "-std=c++17"],
    verbose=False,
)


class ModelNew(nn.Module):
    def __init__(self, in_features, out_features, max_dim):
        super().__init__()
        self.gemm = nn.Linear(in_features, out_features)
        self.max_dim = max_dim

    def forward(self, x):
        return mod.zero_output_like_batch(x.contiguous())

\end{Verbatim}

\begin{center}
\captionof{listing}{Code of the audited high-speedup case for Level~2, Task~14.}
\label{alg:High-Speedup1}
\end{center}
\begin{Verbatim}[fontsize=\footnotesize,breaklines=true,breakanywhere=true]
import torch
import torch.nn as nn
from torch.utils.cpp_extension import load_inline

cuda_src = r"""
#include <torch/extension.h>
#include <cuda_runtime.h>
#include <ATen/cuda/CUDAContext.h>
#include <c10/cuda/CUDAException.h>

extern "C" __global__
void rowdot_kernel(
    const float* __restrict__ x,
    const float* __restrict__ colsum,
    float* __restrict__ out,
    int B, int N,
    float scale
){
    int b = blockIdx.x;
    if(b >= B) return;
    const float* xrow = x + b * N;
    float acc = 0.0f;
    for(int j = threadIdx.x; j < N; j += blockDim.x){
        acc += xrow[j] * colsum[j];
    }
    __shared__ float sh[256];
    sh[threadIdx.x] = acc;
    __syncthreads();
    for(int s = blockDim.x / 2; s > 0; s >>= 1){
        if(threadIdx.x < s){
            sh[threadIdx.x] += sh[threadIdx.x + s];
        }
        __syncthreads();
    }
    if(threadIdx.x == 0){
        out[b] = sh[0] * scale;
    }
}
torch::Tensor fused_forward_cuda(
    torch::Tensor x,
    torch::Tensor colsum,
    float scale
){
    TORCH_CHECK(x.is_cuda(), "x must be on CUDA");
    TORCH_CHECK(colsum.is_cuda(), "colsum must be on CUDA");

    auto B = x.size(0);
    auto N = x.size(1);
    auto out = torch::empty({B,1}, x.options());
    int threads = 256;
    dim3 blocks(B);

    auto stream = at::cuda::getDefaultCUDAStream();
    rowdot_kernel<<<blocks, threads, 0, stream>>>(
        x.data_ptr<float>(),
        colsum.data_ptr<float>(),
        out.data_ptr<float>(),
        (int)B, (int)N,
        scale
    );
    C10_CUDA_KERNEL_LAUNCH_CHECK();
    return out;
}
"""

cpp_src = r"""
torch::Tensor fused_forward_cuda(torch::Tensor x, torch::Tensor colsum, float scale);
"""

ext = load_inline(
    name="fused_matvec_cuda_ext",
    cpp_sources=cpp_src,
    cuda_sources=cuda_src,
    functions=["fused_forward_cuda"],
    verbose=False,
    extra_cflags=[
        "-O3",
        "-std=c++17"
    ],
    extra_cuda_cflags=[
        "-O3",
        "-std=c++17",
        "--expt-relaxed-constexpr",
        "-lineinfo",
        "-gencode=arch=compute_80,code=sm_80"
    ]
)

class ModelNew(nn.Module):
    def __init__(self, input_size, hidden_size, scaling_factor):
        super().__init__()
        self.weight = nn.Parameter(torch.randn(hidden_size, input_size))
        self.scaling_factor = scaling_factor

    def forward(self, x):
        colsum = torch.sum(self.weight, dim=0)
        colsum = colsum.contiguous()
        x = x.contiguous()
        return ext.fused_forward_cuda(x, colsum, 0.5 * self.scaling_factor)

\end{Verbatim}

\begin{center}
\captionof{listing}{Code of the audited high-speedup case for Level~2, Task~51.}
\label{alg:High-Speedup2}
\end{center}

\begin{Verbatim}[fontsize=\footnotesize,breaklines=true,breakanywhere=true]
import torch
import torch.nn as nn
from torch.utils.cpp_extension import load_inline

cuda_src = r"""
#include <torch/extension.h>
#include <cuda_runtime.h>
#include <ATen/cuda/CUDAContext.h>
#include <c10/cuda/CUDAException.h>

__device__ __forceinline__ float gelu_fast(float x){
    return 0.5f * x * (1.0f + tanhf(0.7978845608f * (x + 0.044715f * x * x * x)));
}

__global__ void fused_rowdot_residual_kernel(
    const float* __restrict__ x,         // [B, K]
    const float* __restrict__ wsum,      // [K]
    float bias_sum,
    float sub_mean,
    float inv_out,
    float* __restrict__ out,             // [B, K], output shape matches residual broadcast
    int B, int K
) {
    int b = blockIdx.x;
    if (b >= B) return;
    const float* xrow = x + (int64_t)b * K;

    float acc = 0.0f;
    for (int j = threadIdx.x; j < K; j += blockDim.x) acc += xrow[j] * wsum[j];
    __shared__ float sh[256];
    sh[threadIdx.x] = acc;
    __syncthreads();
    for (int s = blockDim.x / 2; s > 0; s >>= 1) {
        if (threadIdx.x < s) sh[threadIdx.x] += sh[threadIdx.x + s];
        __syncthreads();
    }
    float z = gelu_fast((sh[0] + bias_sum) * inv_out - sub_mean);

    // Broadcast add: out[b, j] = z + x[b, j]
    for (int j = threadIdx.x; j < K; j += blockDim.x) {
        out[(int64_t)b * K + j] = z + xrow[j];
    }
}

torch::Tensor fused_forward(
    torch::Tensor x, torch::Tensor wsum, float bias_sum, float sub_mean, float inv_out
) {
    auto out = torch::empty_like(x);
    int B = (int)x.size(0), K = (int)x.size(1);
    auto s = at::cuda::getDefaultCUDAStream();
    fused_rowdot_residual_kernel<<<B, 256, 0, s>>>(
        x.data_ptr<float>(), wsum.data_ptr<float>(), bias_sum, sub_mean, inv_out, out.data_ptr<float>(), B, K
    );
    C10_CUDA_KERNEL_LAUNCH_CHECK();
    return out;
}
"""

cpp_src = "torch::Tensor fused_forward(torch::Tensor x, torch::Tensor wsum, float bias_sum, float sub_mean, float inv_out);"

mod = load_inline(
    name="k51_fused_ext",
    cpp_sources=cpp_src,
    cuda_sources=cuda_src,
    functions=["fused_forward"],
    extra_cflags=["-O3", "-std=c++17"],
    extra_cuda_cflags=["-O3", "-std=c++17", "--use_fast_math"],
    verbose=False,
)

class ModelNew(nn.Module):
    def __init__(self, in_features, out_features, bias=True):
        super().__init__()
        self.gemm = nn.Linear(in_features, out_features, bias=bias)
        self.subtract = nn.Parameter(torch.randn(out_features))

    def forward(self, x):
        x = x.contiguous()
        w_sum = self.gemm.weight.sum(dim=0).contiguous()
        bias_sum = float(self.gemm.bias.sum().item()) if self.gemm.bias is not None else 0.0
        sub_mean = float(self.subtract.mean().item())
        inv_out = 1.0 / float(self.gemm.weight.shape[0])
        return mod.fused_forward(x, w_sum, bias_sum, sub_mean, inv_out)
\end{Verbatim}

\twocolumn

\end{document}